\documentclass[11pt]{article}

\usepackage[preprint]{acl}

\usepackage{times}  
\usepackage{helvet}  
\usepackage{courier}  

\usepackage{graphicx} 
\usepackage{natbib}   
\usepackage{caption}  
\usepackage{booktabs} % For nice table rules
\usepackage{multirow} % For multi - row cells
\usepackage{graphicx} % For possible graphic operations (used for font size here)
\usepackage{array} % For custom column types (optional, used for centering in p columns)
\usepackage{tabularray}
\usepackage{soul} 
\usepackage{algcompatible}
\usepackage{subcaption}
\usepackage{inconsolata}
\usepackage{algorithm}
\usepackage{colortbl}
\usepackage[noend]{algpseudocode}  % conflict with algorithmic
\usepackage{array} 
\usepackage{url}
\usepackage{etoc}
\usepackage{graphicx}
\usepackage{multirow}
\usepackage{booktabs} 
\usepackage{tcolorbox} 
\usepackage{hyperref}
\usepackage{graphicx}

\usepackage{amsmath, amssymb} 
\allowdisplaybreaks 
\usepackage{amsfonts} 

\usepackage{booktabs} 
\usepackage{graphicx}

\usepackage{url} 
\newcommand{\methodname}{PIR}

\usepackage{tcolorbox}
\tcbuselibrary{skins, breakable}
\usepackage{paracol}
\usepackage{paracol}                
\usepackage{xcolor}                
\usepackage{caption}
\usepackage{cuted}

\usepackage{newfloat}
\usepackage{listings}
\DeclareCaptionStyle{ruled}{labelfont=normalfont,labelsep=colon,strut=off}
\floatstyle{ruled}
\newfloat{listing}{tb}{lst}{}
\floatname{listing}{Listing}

\title{Precedent-Informed Reasoning: Mitigating Overthinking in Large Reasoning Models via Test-Time Precedent Learning}

\newtcolorbox[auto counter]{mybox}[1][]{
  title=My Box \thetcbcounter,
  label=mybox:\thetcbcounter,
  #1
}
\lstdefinelanguage{json}{
  basicstyle=\ttfamily\small,
  numberstyle=\scriptsize\color{gray},
  stringstyle=\color{blue},
  keywordstyle=\color{teal},
  commentstyle=\color{gray}\itshape,
  morestring=[b]",
  literate=
    *{0}{{{\color{teal}0}}}{1}
     {1}{{{\color{teal}1}}}{1}
     {2}{{{\color{teal}2}}}{1}
     {3}{{{\color{teal}3}}}{1}
     {4}{{{\color{teal}4}}}{1}
     {5}{{{\color{teal}5}}}{1}
     {6}{{{\color{teal}6}}}{1}
     {7}{{{\color{teal}7}}}{1}
     {8}{{{\color{teal}8}}}{1}
     {9}{{{\color{teal}9}}}{1}
     {:}{{{\color{black}{:}}}}{1}
     {,}{{{\color{black}{,}}}}{1}
     {\{}{{{\color{black}{\{}}}}{1}
     {\}}{{{\color{black}{\}}}}}{1}
     {[}{{{\color{black}{[}}}}{1}
     {]}{{{\color{black}{]}}}}{1}
}

\author{
    Qianyue Wang \textsuperscript{\rm 1 \rm 2}~
    Jinwu Hu \textsuperscript{\rm 1 \rm 2}~
    Huanxiang Lin \textsuperscript{\rm 1 }~
    Bolin Chen \textsuperscript{\rm 1 }~\\
    \textbf{Zhiquan Wen} \textsuperscript{\rm 1 \rm 2}~
    \textbf{Yaofo Chen} \textsuperscript{\rm 1 \rm 2}~
    \textbf{Yu Rong} \textsuperscript{\rm 3}~
    \textbf{Mingkui Tan} \textsuperscript{\rm 1} \footnotemark[2] ~\\ 
    \textsuperscript{\scriptsize{\rm 1}}\small{South China University of Technology,}~
    \textsuperscript{\rm 2}\small{Pazhou Laboratory,}~
    \textsuperscript{\rm 3}\small{DAMO Academy, Alibaba Group}~ \\
}

\begin{document}
\maketitle
\footnotetext[1]{Corresponding author: mingkuitan@scut.edu.cn.}

\begin{abstract}
Reasoning in Large Language Models (LLMs) often suffers from inefficient long chain-of-thought traces with redundant self-exploration and validation, {which inflate computational costs and even degrade performance.} Inspired by human reasoning patterns where people solve new problems by leveraging past related cases to constrain search spaces and reduce trial-and-error, we propose \textbf{P}recedent \textbf{I}nformed \textbf{R}easoning (\textbf{\methodname{}}) transforming LRMs’ reasoning paradigm from exhaustive self-exploration to guided learning from precedents.~\methodname{} addresses two key challenges: what precedents to adopt and how to utilize them. First, \textbf{Adaptive Precedent Selection (APS)} constructs, for each question and LRM, a compact set of precedents that are both semantically related and {informative} for the model. It ranks examples by a joint score with semantic similarity and model perplexity, then adapts the amount of precedents to maximize perplexity reduction. Second, \textbf{Test-time Experience Internalization (TEI)} is treated as the test-time learning on precedent-informed instruction, updating lightweight adapters to internalize solution patterns and use them as a prior during subsequent reasoning. Experiments across mathematical reasoning, scientific Q\&A, and code generation demonstrate that~\methodname{} consistently shortens reasoning traces while maintaining or improving final accuracy across LLMs, yielding outstanding accuracy-efficiency trade-offs.

\end{abstract}
\section{Introduction}

Large Language Models (LLMs)~\cite{openai23gpt, touvron2023llama} have demonstrated strong performance on a range of natural language processing tasks, including question answering~\cite{EUP} and text generation~\cite{icl_wangqianyue}.
However, standard LLMs exhibit degraded performance in complex reasoning scenarios that require multi-step deduction and coherent integration of intermediate conclusions.
To address this limitation, Large Reasoning Models (LRMs)~\cite{stop_oevrthinking_survey}, such as OpenAI’s o1~\cite{openai_o1} and DeepSeek-R1~\cite{DeepSeek-R1}, have been developed to decompose complex problems into step-by-step reasoning, leading to substantial improvements on reasoning-intensive tasks~\cite{math500,gsm8k,numina_math_datasets}.

{Unfortunately, LRMs often engage in redundant self-exploration~\cite{redundant_explor} and validation~\cite{redundant_self} in reasoning-intensive tasks, a phenomenon referred to as \textbf{overthinking}~\cite{stop_oevrthinking_survey,chennot}. This redundancy increases unnecessary token generation and computation, which raises reasoning cost and even hurts accuracy when the model over-commits to exploration.}

{Recently, numerous methods have been proposed to mitigate overthinking in LRMs, which can be broadly categorized into two types. \textbf{Multi-model collaboration} methods~\cite{specutive_thinking,specreason,hu2025fasts} decompose reasoning processes by difficulty and distribute them across models of different scales, reducing the number of generated tokens for complex reasoning tasks. \textbf{Single-model optimization} methods focus on improving inference efficiency within a single model, through high-cost retraining~\cite{s1} or via test-time control strategies such as dynamic early stopping~\cite{early_stop1} and reasoning compression~\cite{reasoning_compress}. While these methods are effective in reducing computational cost or inference latency, they largely operate by controlling the length or budget of the reasoning trace, without altering the underlying trial-and-error-based reasoning. As a result, each question is still treated as an independent reasoning instance, and LRMs are not explicitly encouraged to reuse solution patterns across questions.}

\begin{figure}[!t]
    \centering
    \includegraphics[width=0.46\textwidth]{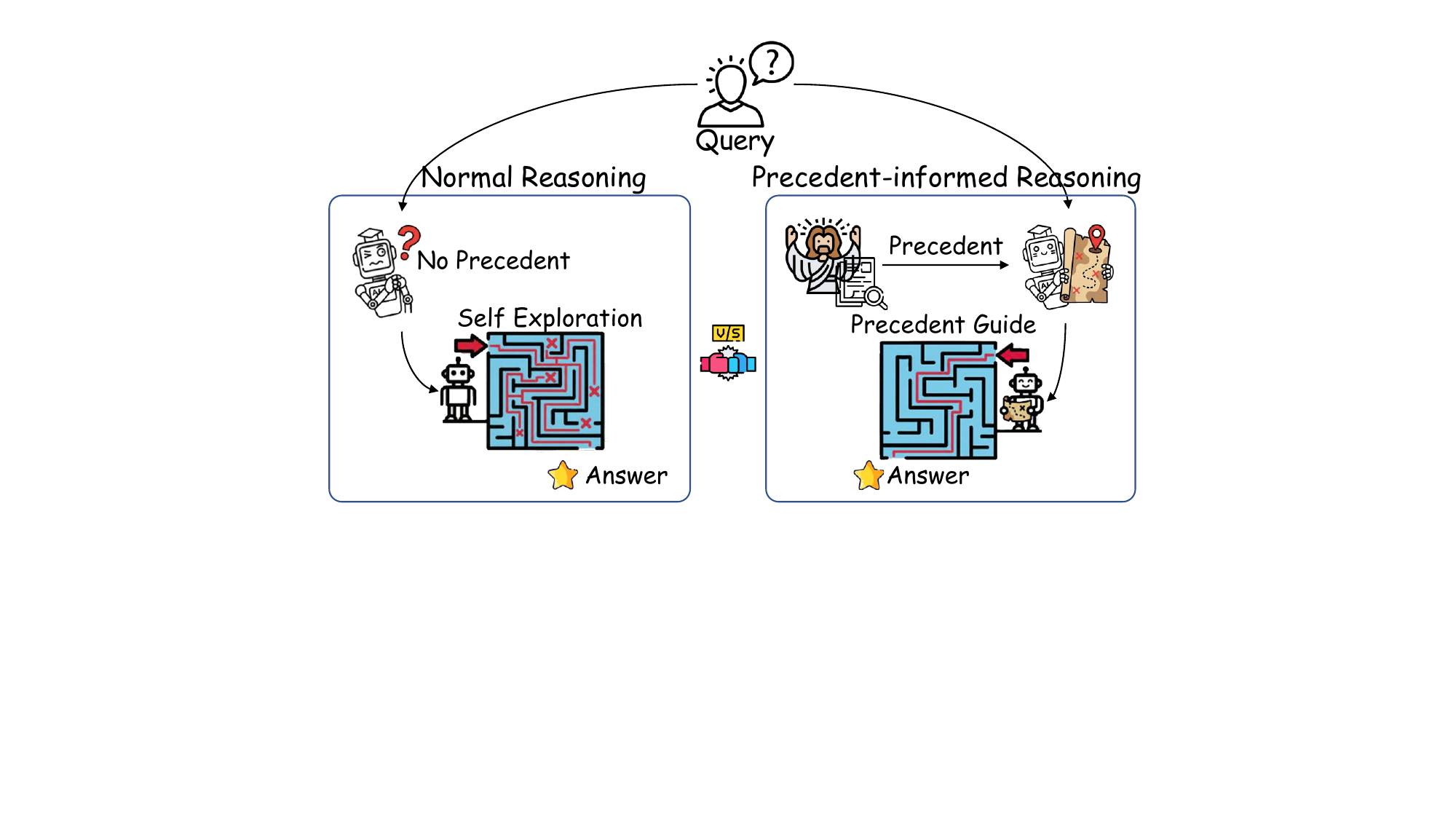}
    \caption{The illustration of different reasoning paradigms. Unlike normal reasoning that relies on self-exploration, precedent-informed reasoning leverages precedent experience to reduce the search space, improving reasoning efficiency.}
    \vspace{-10pt}
    \label{fig: compare}
\end{figure}

{In contrast, when solving complex reasoning problems, humans adopt a \textit{precedent-based reasoning style}, recalling or retrieving a set of similar precedents and imitating their solution patterns~\cite{polya1945solve} {to constrain the solution search space, reducing trial-and-error exploration.} As illustrated in Figure~\ref{fig: compare}, we aim to seek a reasoning paradigm for LRMs that aligns with human precedent-based reasoning, enabling LRMs to solve problems by imitating the solution patterns in precedents instead of relying solely on self-exploration. {However, enabling precedent-based reasoning in LRMs raises two challenges: \textit{(i) What to imitate?} Given a question and a target LRM, how to select a set of precedents that are both semantically related and informative to guide the model’s reasoning. \textit{(ii) How to imitate?} We need a mechanism that enables the model to actively internalize solution patterns from precedents and use them as a prior for subsequent reasoning.
}}

{In this work, we propose \textbf{P}recedent \textbf{I}nformed \textbf{R}easoning (\textbf{\methodname{}}) at test time to mitigate overthinking by transforming the reasoning process of LRMs from redundant self-exploration to precedent internalization and precedent-guided reasoning. Specifically, we introduce \textbf{A}daptive \textbf{P}recedent \textbf{S}election (\textbf{APS}), which combines semantic similarity and model perplexity to adaptively determine a compact set of informative precedents for each question. We further propose \textbf{T}est-time \textbf{E}xperience \textbf{I}nternalization (\textbf{TEI}), which treats precedent-guided reasoning as test-time learning on a precedent-informed instruction constructed from the selected precedents and the target question, enabling the model to internalize their solution patterns {and reason on a reduced solution search space to improve reasoning efficiency.} Main contributions are as follows:}

\begin{itemize}
    \item {\textbf{From Self-Exploration to Precedent-informed Reasoning Paradigm.} We introduce a precedent-based reasoning paradigm that mitigates the overthinking phenomenon by replacing redundant self-exploration and validation with precedent internalization, constraining LRMs to a smaller, strategy-guided search space.}
    \item {\textbf{\methodname{}: Test-time Precedent Selection and Experience Internalization.} We propose precedent-informed Reasoning, a test-time method that i) uses Adaptive Precedent Selection to choose a compact set of precedents and ii) applies Test-time Experience Internalization to internalize the solution patterns in precedents via label-free test-time learning on precedent-informed instructions.} 
    \item {\textbf{Empirical accuracy–efficiency trade-offs across tasks and models.} We demonstrate on mathematical reasoning, scientific question answering, and code generation benchmarks across multiple LRMs that our method consistently shortens reasoning traces and reduces computation while matching or improving accuracy, achieving strong accuracy–efficiency trade-offs.}
\end{itemize}

\section{Related Work}
\label{sup: related work}
\subsection{Efficient Reasoning for LRMs}
{Efficient reasoning is critical for balancing model performance and computational cost, and existing approaches to improving the efficiency of LRMs can be broadly grouped into two categories. \textbf{Multi-model collaboration} methods exploit the complementarity of models with different capacities for complex reasoning tasks.} {
SpecReason~\cite{specreason} delegates part of the base model's reasoning tasks to a smaller model for speculative execution. Speculative Thinking~\cite{specutive_thinking} primarily uses a smaller model for reasoning, with a larger model providing intervention and correction as needed.} While effective to some extent, these methods require {considerable} GPU resources to support the simultaneous loading of models with different scales, which limits their practical usability.
{\textbf{Single-Model Optimization} methods aim to improve reasoning efficiency within a single model, either through training-time refinement or test-time control.} Training strategies such as L1~\citep{l1} and s1~\citep{s1} modify reasoning patterns through extensive fine-tuning to reduce redundancy. {Test-time methods include dynamic early exiting~\cite{early_stop1} like DEER~\cite{deer}), which uses pivotal tokens to control output length via a single heuristic without considering the intrinsic limitation of reasoning behavior in LRMs and reasoning compression~\cite{reasoning_compress} like SEAL~\cite{seal}), which compresses reasoning trajectories by adjusting the internal representations of LRMs.}

\subsection{Test time In-context Learning}

{In-context learning (ICL) allows LLMs to leverage information explicitly provided in the prompt rather than relying solely on their imperfect parametric memory~\cite{icl}. ICL improves performance within few references but the improvement is limited by prompt formatting, example ordering, and context length constraints~\cite{icl+ft}. Test-time learning takes an active parameter fine-tuning approach that updates the model on a small batch of test examples to adapt to the input distribution, and is proven to be effective in a variety of settings~\cite{ttlforcontextual, ttt1,ttt2,ttt3}. TLM~\cite{PPL_min_learn} shows that using perplexity as the adaptation objective and adapting solely to the input content, without requiring task labels, can yield improvements in task performance. }

\begin{algorithm}[t]
\footnotesize
\caption{The pipeline of~\methodname.}
\label{alg: method}
\begin{algorithmic}[1]
  \Require Question $q$, solved precedent set $E_s = \{(q_i, a_i)\}_{i=1}^N$, model $M$ with parameters $\Theta$, pre-selection size $k$, max reference budget $p$, instruction template $I$.
  \Ensure Reasoning result $y$

  \State \textbf{// Reference example pre-selection}
  \State Construct $E_k \subset E_s$ by selecting the $k$ examples with the largest semantic similarity $s_i$ between $q$ and each $q_i$, where $s_i$ is computed via Eq.~(\ref{eq: ss}).
  \State Construct $E_p \subset E_k$ by selecting the $p$ examples with the largest Reference Score $\mathrm{RS}_i$ for $(q_i, a_i) \in E_k$, where $\mathrm{RS}_i$ is computed from the question perplexity of $q_i$ under $M_\Theta$ via Eq.~(\ref{eq: rs}).

  \State \textbf{// Adaptive number of reference examples}
  \State Construct the question-only instruction $I_0 = I(q)$ whose perplexity $P_0 = P(I_0; \Theta)$ is computed under $\Theta$.
  \For{$m = 1,\ldots,p$}
      \State Construct the precedent-informed instruction $I_m = I\big(q, \{(q_{(j)}, a_{(j)})\}_{j=1}^m\big)$ using the first $m$ examples in $E_p$, whose perplexity $P(I_m; \Theta)$ under parameters $\Theta$ yields the relative reduction $\Delta P(I_m, \Theta) = P_0 - P(I_m; \Theta)$.
  \EndFor
  \State Choose $m^\star$ such that $\Delta P(I_{m^\star}, \Theta) \ge \Delta P(I_m, \Theta)$ for all $m = 1,\ldots,p$.
  \State Set $E_0^\star(q) = \{(q_{(j)}, a_{(j)})\}_{j=1}^{m^\star}$.
  \State Set $I^\star = I_{m^\star}$.

  \State \textbf{// Test-time instruction adaptation for solution internalization}
  \State Initialize LoRA parameters $\Delta \Theta = BA$ by sampling $A$ from a Gaussian distribution, and $B$ is initialized to $0$.
  \State Update $\Delta \Theta$ by one backward pass to minimize $P(I^\star; \Theta + \Delta \Theta)$ as in Eq.~(\ref{eq: min_obj_aim}).
  \State Obtain adapted parameters $\Theta^\star = \Theta + \Delta \Theta$.
  \State \textbf{// Reasoning for final answer}
  \State Generate the reasoning process and final answer $y$ for $q$ using $M$ with parameters $\Theta^\star$.
\end{algorithmic}
\end{algorithm}

\begin{figure*}[t]
    \centering
    \begin{subfigure}[b]{0.32\textwidth}
        \centering
        \includegraphics[width=\linewidth,height=0.5\textheight,keepaspectratio,trim=7 7 5 5,clip]{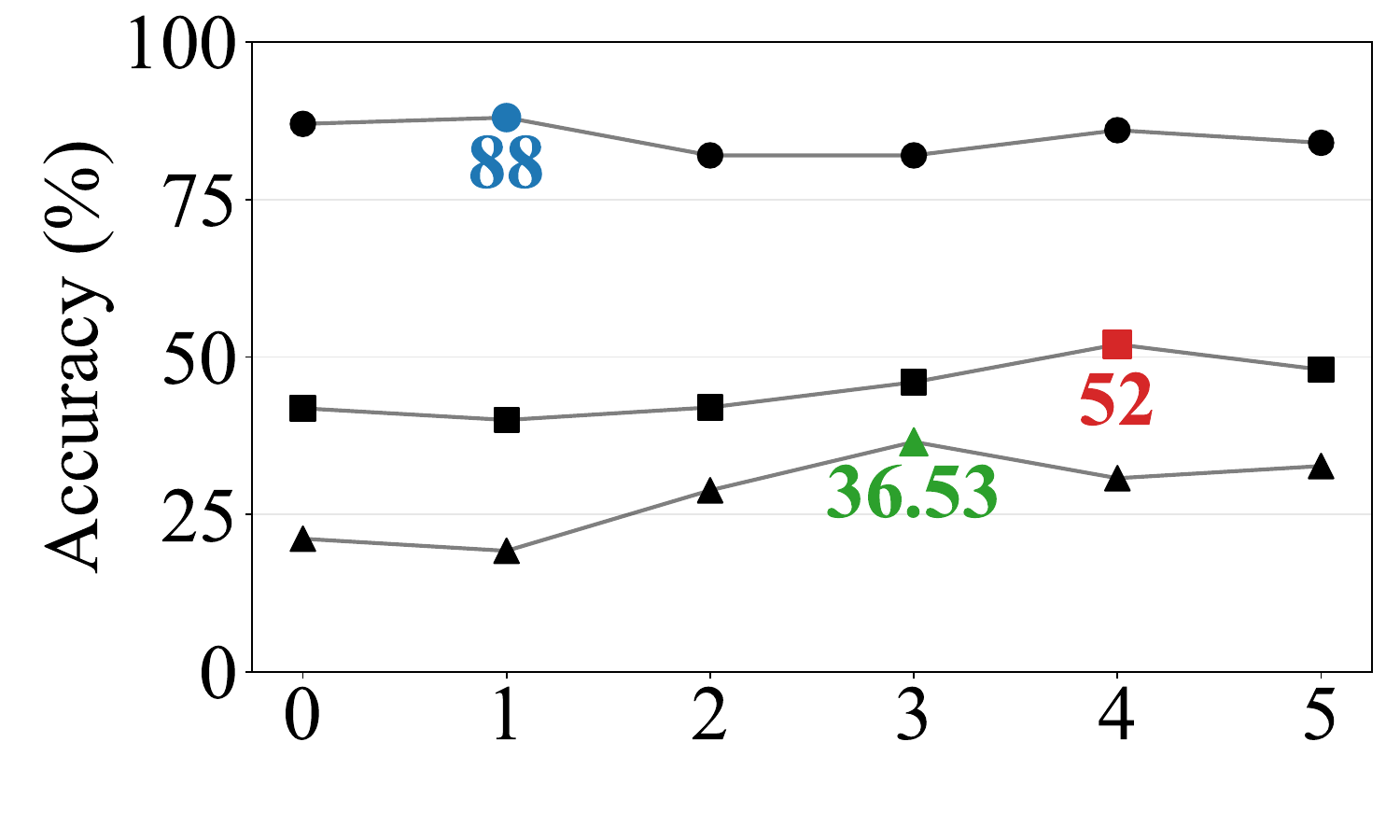}
        \vspace{2pt}
        \includegraphics[width=\linewidth,height=0.5\textheight,keepaspectratio,trim=7 0 5 5,clip]{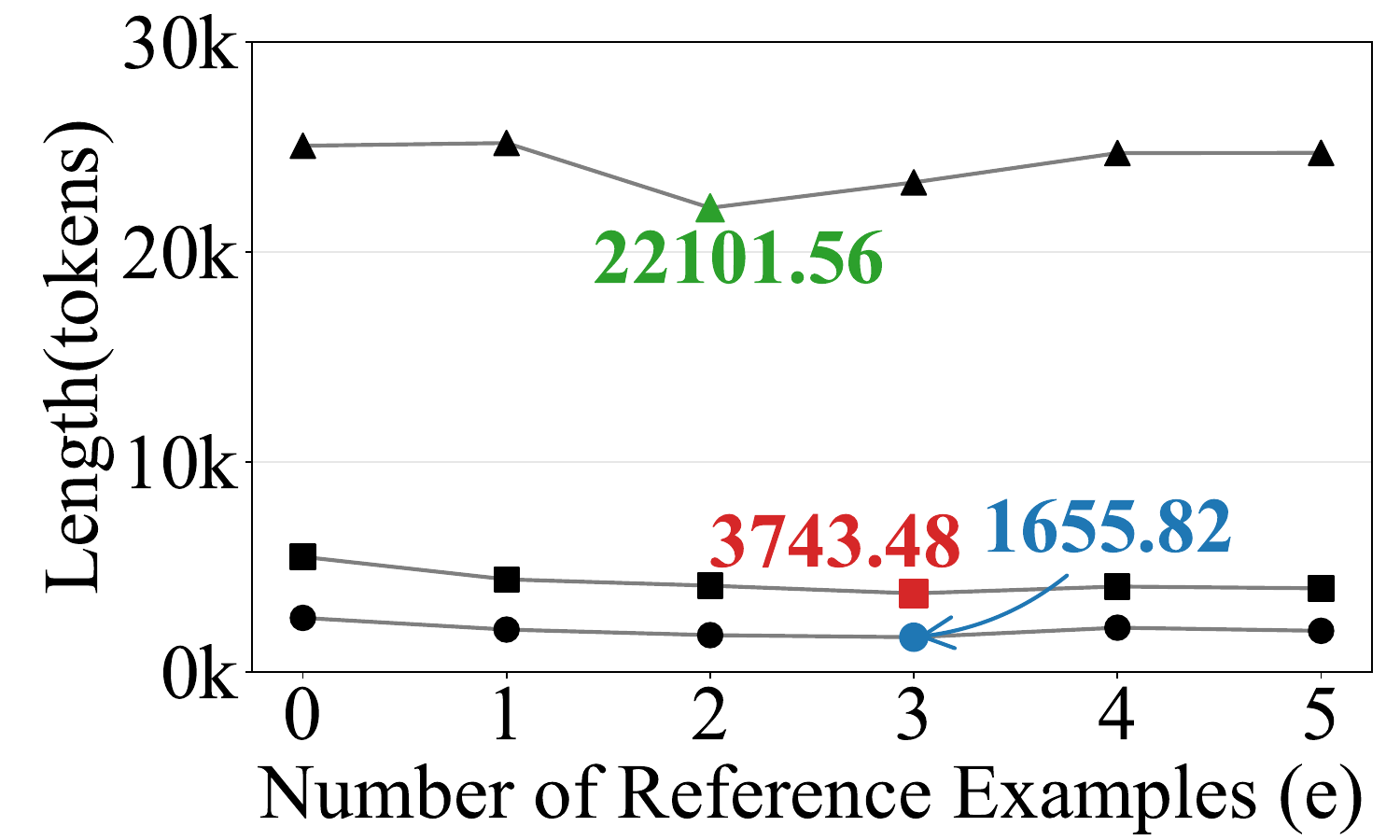}
        \caption{DeepSeek-R1-DQ-7B}
        \label{subfig:a}
    \end{subfigure}
    \hfill
    \begin{subfigure}[b]{0.32\textwidth}
        \centering
        \includegraphics[width=\linewidth,height=0.5\textheight,keepaspectratio,trim=7 7 5 5,clip]{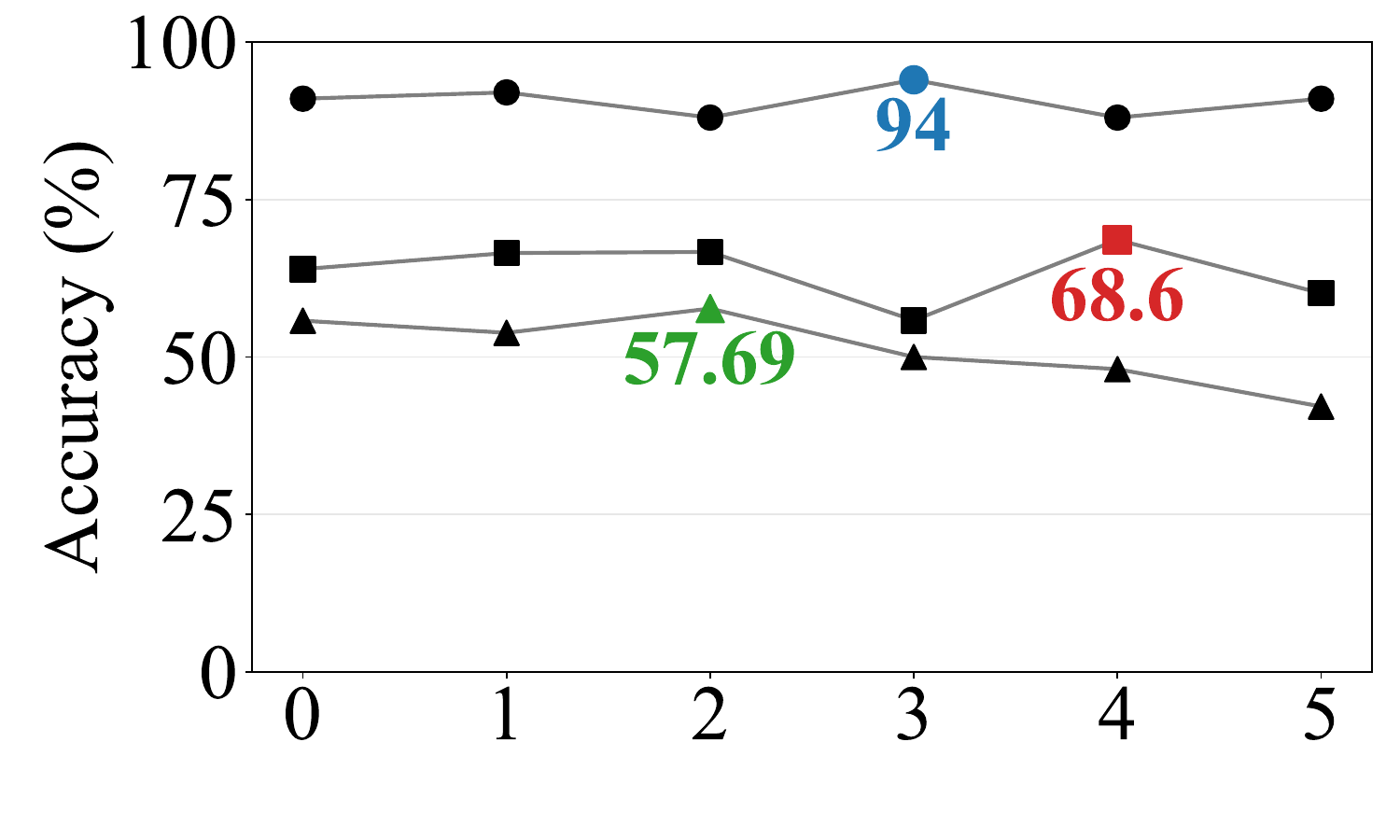}
        \vspace{2pt}
        \includegraphics[width=\linewidth,height=0.5\textheight,keepaspectratio,trim=7 0 5 5,clip]{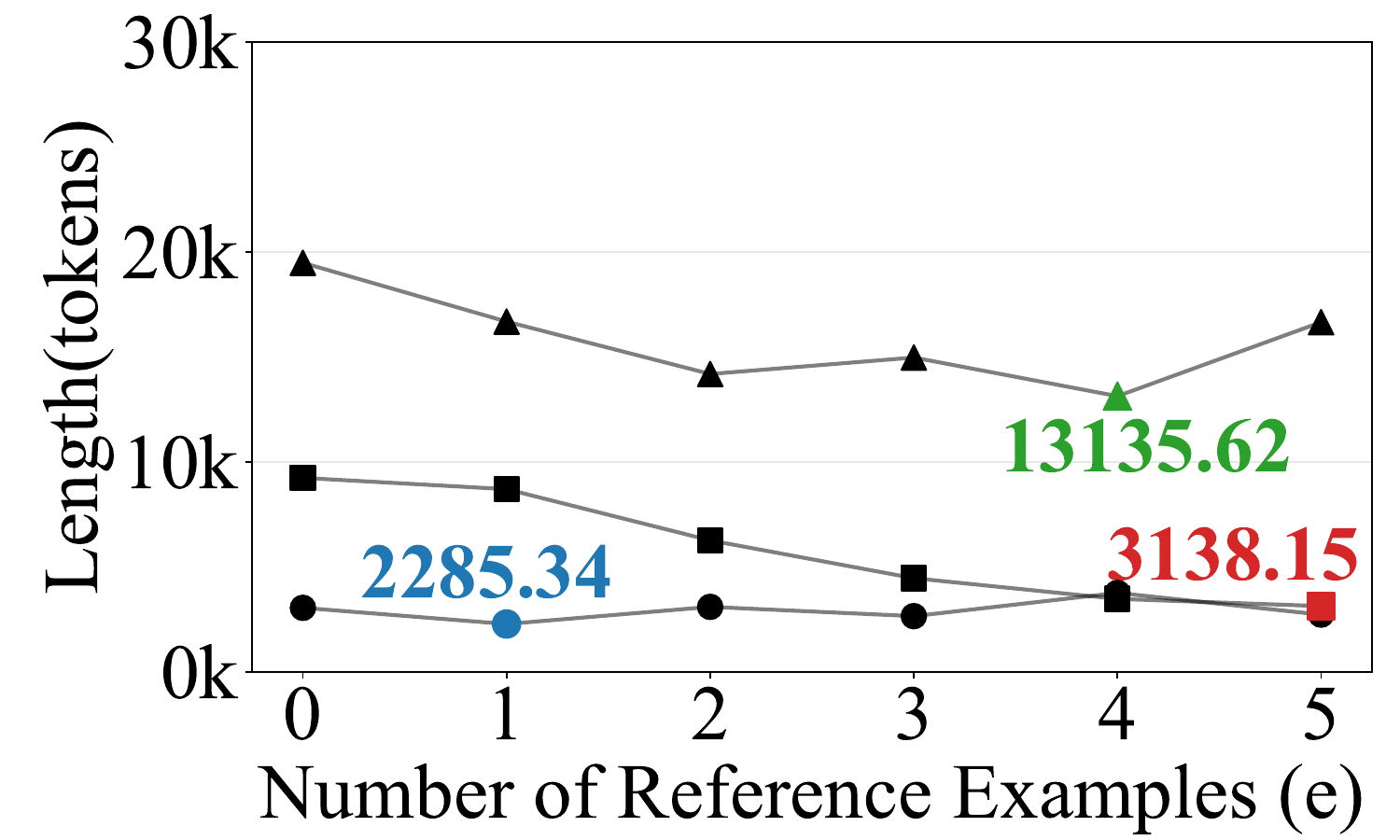}
        \caption{DeepSeek-R1-DQ-32B}
        \label{subfig:b}
    \end{subfigure}
    \hfill
    \begin{subfigure}[b]{0.32\textwidth}
        \centering
        \includegraphics[width=\linewidth,height=0.5\textheight,keepaspectratio,trim=7 7 5 5,clip]{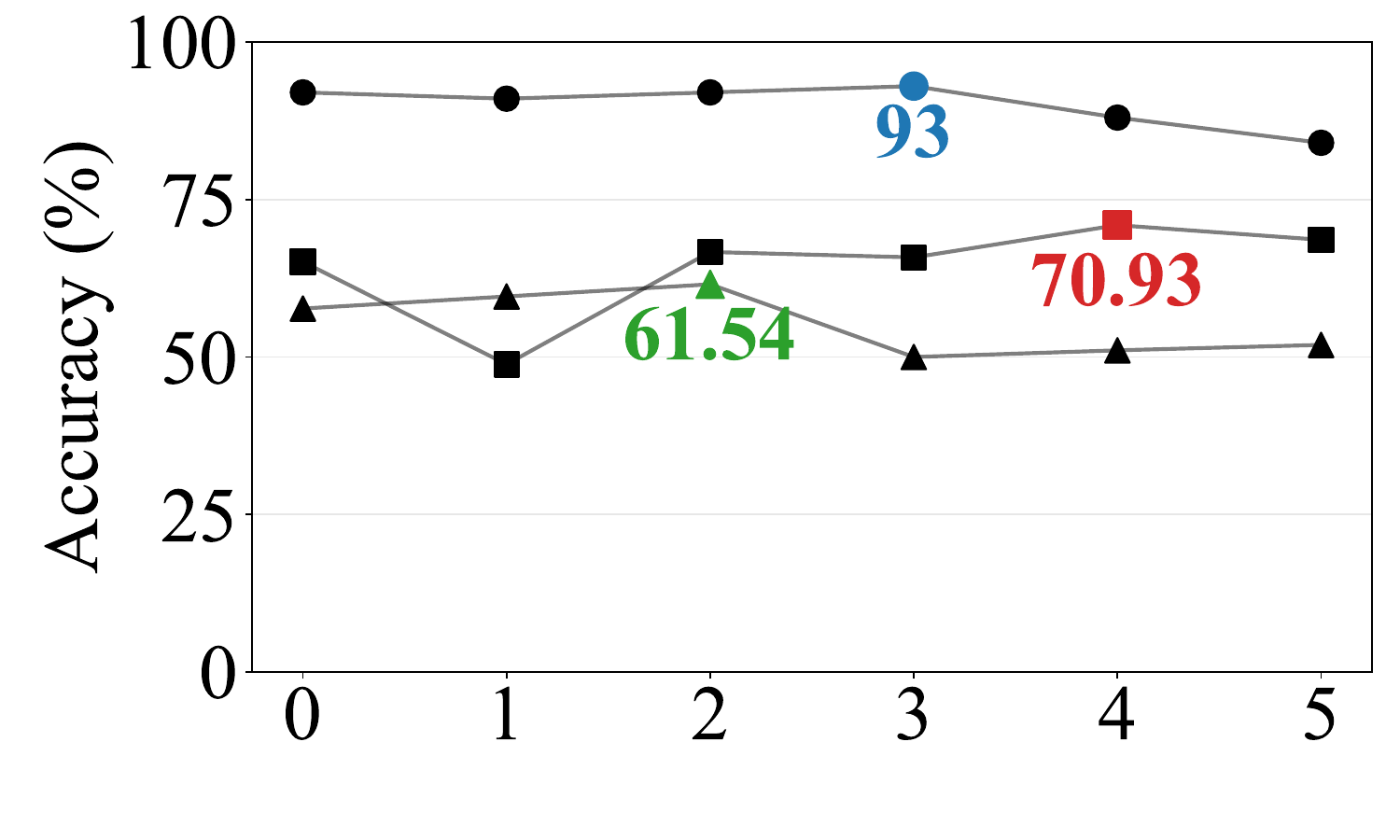}
        \vspace{2pt}
        \includegraphics[width=\linewidth,height=0.5\textheight,keepaspectratio,trim=7 0 5 5,clip]{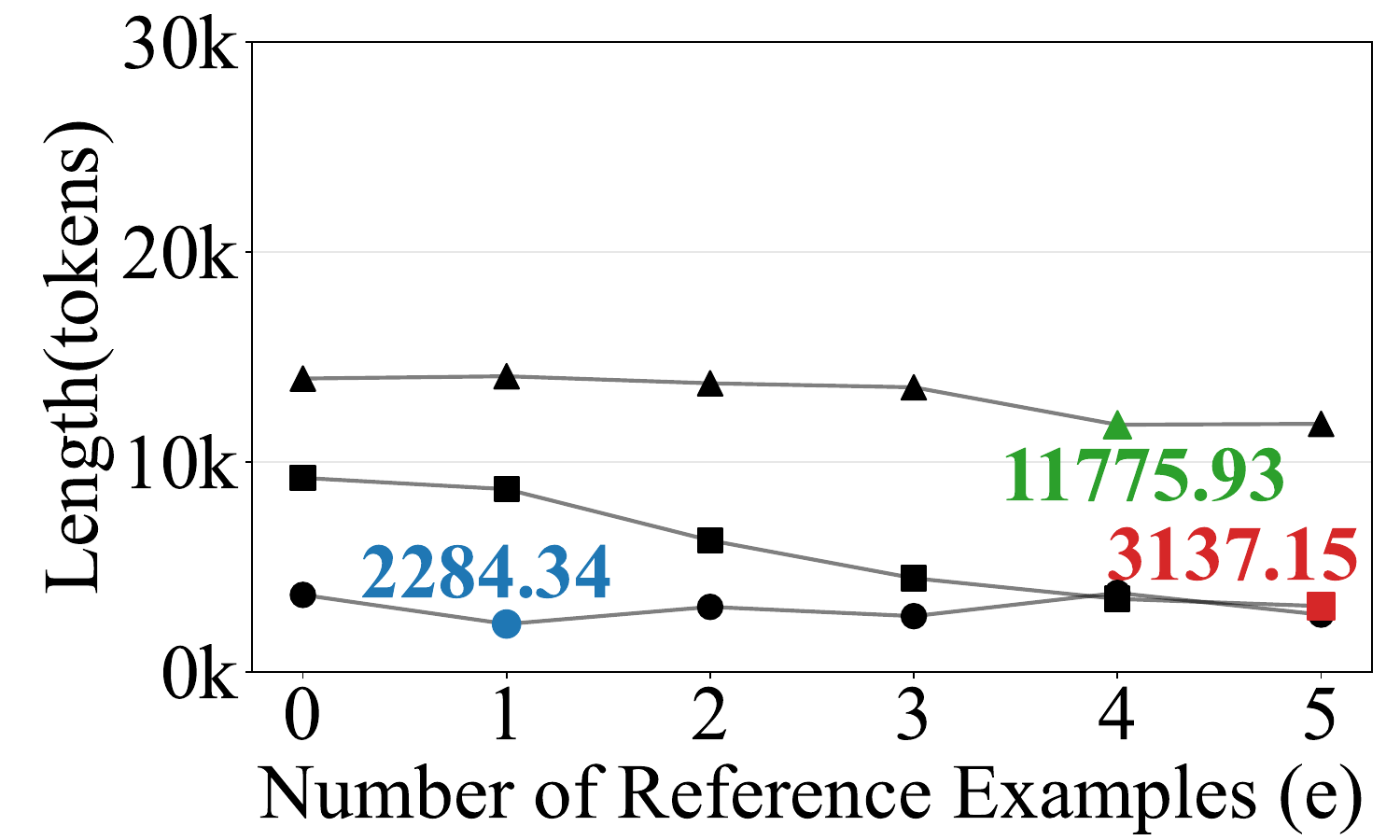}
        \caption{QwQ-32B}
        \label{subfig:c}
    \end{subfigure}
    \caption{{The accuracy and length on DeepSeek-R1-Qwen2.5-7B/32B and QwQ-32B as incrementally vary the number of $RS$-filtered precedents in the reasoning prompt. Best accuracy gains are achieved within the given reference budget. We evaluate 100 randomly sampled instances per dataset. Markers denote the best result for each model-dataset pair: $\textcolor[rgb]{0.122,0.467,0.706}{\bullet}$ as MATH500, $\textcolor[rgb]{0.839,0.153,0.157}{\blacksquare}$ as GPQA Diamond, $\textcolor[rgb]{0.173,0.616,0.173}{\blacktriangle}$ as LiveCode Bench. See Appendix~\ref{sup: pre-exp} for details.}}
    \label{fig: pre_exp}
    \vspace{-10pt}

\end{figure*}

\section{Problem Statement and Motivation}

\textbf{Problem statement.} Given a question $q$, we use a reasoning-enhanced model $M_\theta$ to produce an answer, and at test time we have an additional set of solved precedents $E_s = \{(q_i,a_i)\}_{i=1}^N$. For each $q$, we select a reference set $E_s^\star(q)$ from an additional precedent set $ E_s$ and apply lightweight test-time parameter updates, obtaining an adapted model $M_\theta^\star$ conditioned on $q$ and its references. The adapted model $M_\theta^\star$ then generates a reasoning trajectory and final answer for $q$ by imitating the reasoning patterns in $E_s^\star(q)$, with the goal of reducing reasoning length and computational cost while preserving or improving task accuracy relative to the base model $M_\theta$.

{\noindent\textbf{Motivation.} Humans often solve hard problems by recalling a few precedents, abstracting reusable strategies, and applying them to new instances, especially in legal case judgment~\cite{legal}, disease diagnosis~\cite{diagnosis}, and complex mathematical problem solving~\cite{polya1945solve}, which narrows the search space and reduces trial-and-error exploration. In contrast, LRMs are not explicitly encouraged to internalize and reuse solution patterns across questions and tackle each query in isolation with much exploration and validation for solution, leading to overthinking with increased computational cost and potential performance degradation. The gap between precedent-based reasoning and model behavior motivates us to develop a method that enables LRMs to reason based on actively imimitating and internalizing the solution patterns from precedents, steering them away from purely self-exploraton and reducing redundancy while maintaining task performance.}

\section{Precedent-informed Reasoning}

{In this paper, we propose \textbf{P}recedent \textbf{I}mitation \textbf{R}easoning (\textbf{\methodname{}}), a precedent-guided test-time adaptation framework that mitigates overthinking by shifting LRMs from exhaustive self-exploration and self-validation to precedent-informed reasoning. Our design is organized around two central questions: \textit{(i) what to imitate?} and \textit{(ii) how to imitate?} For \textit{what to imitate}, Adaptive Precedent Selection (APS) selects, for each question and target model, a compact set of precedents with an adaptive total amount by semantic similarity to the target question and 
{model-specific} informativeness measured by perplexity. For \textit{how to imitate}, Test-time Experience Internalization (TEI) performs test-time adaptation on the LoRA module to a precedent-informed instruction built from the selected precedents so that the model internalizes their solution patterns and uses them as a behavioral prior during subsequent reasoning, thereby reducing redundant exploration and verification. The pseudo-code of~\methodname{} is in Algorithm~\ref{alg: method}.}

\subsection{Adaptive Precedent Selection}

{To address \textit{what to imitate}, we propose \textbf{A}daptive \textbf{P}recedent \textbf{S}election (\textbf{APS}), which constructs a compact precedent set $E^\star(q)$ from external precedent set $E_s$ for each question $q$ and target model $M_\Theta$ pair. APS first performs coarse filtering by ranking candidates with a joint score of question-level semantic similarity and model perplexity. Inspired by the idea of prequential evaluation~\cite{icl+ft}, which determines hyperparameters by considering the downstream performance, we progressively add one from top-ranked precedents together with the given question to form precedent-informed instructions and determine the precedent size that yields the largest marginal perplexity reduction for the instruction, which reflects the benefit of subsequent test-time adaptation~\cite{PPL_min_learn}. We detail its two stages, starting with multi-criteria pre-selection and followed by task-aware budget determination.}

{\textbf{Top-$k$ Semantic Similarity Selection.}
Given a target question $q$ and a precedent set
$E_s=\{(q_i,a_i)\}_{i=1}^N$, we measure the semantic relevance between $q$ and each
precedent question $q_i$ in an embedding space:
\begin{equation}
s_i = \cos\!\big(\phi(q), \phi(q_i)\big), \quad (q_i,a_i)\in E_s,
\label{eq: ss}
\end{equation}
where $\phi(q)$ and $\phi(q_i)$ are the embeddings of the
target question and the precedent question. $E_k$ is formed by the top-$k$ examples with the highest $s_i$.}

\textbf{Top-$p$ Reference Score Selection.}
{For each $(q_i,a_i)\in E_k$, we further assess how familiar it is to the target reasoning model $M_\Theta$ by the perplexity of its question tokens $\{x_1,\dots,x_{|q_i|}\}$:}
\begin{equation}
\mathcal{P}_i = \exp\!\left(-\frac{1}{|q_i|}\sum_{t=1}^{|q_i|} \log p_{M_\Theta}(x_t \mid x_{<t})\right).
\label{eq: ppl}
\end{equation}
Let $\mathcal{P}$ denote the set of $\mathcal{P}_i$. {Since perplexity values can vary by orders of magnitude across models and are on a different scale from cosine similarity scores, we apply min–max normalization to $\mathcal{P}_i$:
\begin{equation}
    \widehat{\mathcal{P}}_i = \dfrac{\max(\mathcal{P}) - \mathcal{P}_i}{\max(\mathcal{P}) - \min(\mathcal{P})}, %\, e_i \in E_k.
\end{equation}
We then define the Reference Score as:
\begin{equation}
\label{eq: rs}
    \text{RS}_i = s_i + \widehat{\mathcal{P}}_i.
\end{equation}
}
{This score prioritizes precedents that are both semantically relevant to $q$ and informative for $M_\Theta$ to imitate. We then take the Top-$p$ precedents by the highest $\text{RS}_i$ to form $E_p$. Next, we decide how many precedents to use from $E_p$ to best support reasoning and we conduct a preliminary study, leading to the following observation:
}

\textbf{Observation 1: No universal optimal precedent quantity.}
From Figure~\ref{fig: pre_exp}, the best number of precedents for accuracy varies across model-dataset pairs, reflecting differences in models' context utilization as well as variation in task difficulty. 
This motivates an adaptive determination of the number of precedents.

{Based on \textbf{Observation 1}, we design an adaptive quantity determination mechanism inspired by prequential evaluation~\cite{icl+ft} for the downstream task. {We select the reference budget that maximizes the perplexity reduction of the precedent-informed instruction (see Appendix~\ref{sup: ins_template}), which benefits subsequent adaptation~\cite{PPL_min_learn}.} We sort $E_p$ in descending order of $\text{RS}_i$ and construct a sequence of precedent-informed instructions $\{I_m\}_{m=1}^p$, where $I_m$ contains the target question and the first $m$ examples. For each $I_m$, we compute the relative perplexity reduction $\Delta P(I_m,\Theta)$, compared with the question-only instruction and choose:}
\begin{equation}
m^\star = \arg\max_{m \in [1,p]} \bigl(-\Delta P(I_m,\Theta)\bigr).
\label{eq: dynamic_m}
\end{equation}
The example subset $E_0^\star$, composed of the top $m^\star$ examples from $E_p$, is used in the next stage of~\methodname{}.

\subsection{{Test-time Experience Internalization}}

{\textbf{Observation 2: Examples alone do not guarantee efficient reasoning.} Figure~\ref{fig: pre_exp} shows that the precedent quantity that maximizes accuracy does not necessarily minimize reasoning length. This suggests that appending precedents as passive in-context is insufficient, motivating a mechanism that enables the model to adapt to the selected precedents and actively use them as reasoning prior, instead of trial-and-error self-exploration.}

{Based on \textbf{Observation 2}, we introduce Test-time Experience Internalization (TEI), which treats the change of reasoning behavior 
as a test-time learning task to instructions for precedent-informed reasoning.
For each question $q$, we construct a precedent-informed instruction $I_{m^\star}$ that concatenates $q$ with its selected reference precedents $E_0^\star$. Instead of expensive offline supervised fine-tuning that requires additional labeled data for every target model, we keep the base parameters $\Theta$ frozen and update only lightweight LoRA~\cite{lora} with parameters $\Delta\Theta$ by minimizing the model perplexity on $I_{m^\star}$, which yields an instance-wise adaptation without task labels~\cite{PPL_min_learn}. TEI optimizes the objective as:
\begin{equation}
\min_{\Delta\Theta} \mathcal{P}(I_{m^\star}; \Theta + \Delta\Theta),
\label{eq: min_obj_aim}
\end{equation}
where $\mathcal{P}(\cdot; \Theta + \Delta\Theta)$ denotes the perplexity under the adapted model. This adaptation results in an adapted model $M^\star$ with parameters $\Theta + \Delta\Theta$. $M^\star$ generates the reasoning and final answer for $q$ under the behavioral prior encoded by $I_{m^\star}$.}

{By enforcing low perplexity on precedent-informed instructions, TEI encourages the model to adapt to internalize solution patterns in precedents and constrain reasoning in a smaller search space~\cite{PPL_min_learn}, reducing redundant exploration and maintaining task performance.}

\section{Experiment}
\begin{table*}[!ht]
  \centering
  \caption{Accuracy ($\mathbf{acc.}$, \%) and average thinking length ($\mathbf{token.}$) across methods and datasets. Best and second-best results are in \textbf{bold} and \underline{underlined}. All methods including~\methodname{} use DeepSeek-R1-32B as the backbone model.}
  \label{tab: com_exp}
  \fontsize{9}{10}\selectfont 
  \setlength{\tabcolsep}{2.5pt}
  \begin{tabular}{l *{10}{c}}
    \toprule
    Method & \multicolumn{2}{c}{GSM8K} & \multicolumn{2}{c}{MATH500} & \multicolumn{2}{c}{AMC AIME} & \multicolumn{2}{c}{GPQA Diamond} & \multicolumn{2}{c}{LiveCodeBench} \\
    \cmidrule(lr){2-3} \cmidrule(lr){4-5} \cmidrule(lr){6-7} \cmidrule(lr){8-9} \cmidrule(lr){10-11}
    & $\mathbf{acc.}$ & $\mathbf{token.}$ & $\mathbf{acc.}$ & $\mathbf{token.}$ & $\mathbf{acc.}$ & $\mathbf{token.}$ & $\mathbf{acc.}$ & $\mathbf{token.}$ & $\mathbf{acc.}$ & $\mathbf{token.}$ \\
    \midrule
    DeepSeek-R1-7B     & 92.60 & 380.24 & 87.00 & 2576.49 & 58.33 & 3858.16 & 22.80 & 5020.86 & 32.68 & 5775.04 \\
    DeepSeek-R1-14B      & 92.60 & 265.91 & 86.00 & 2408.36 & 57.76 & 3523.09 & 24.66 & 4411.01 & 36.57 & 5432.53 \\
    DeepSeek-R1-32B    & 94.20 & 263.04 & 86.80 & 2269.37 & 59.92 & 3661.52 & 36.30 & 5028.59 & 41.02 & 5224.52 \\
    QwQ-32B~\cite{qwq32b}              & 93.60 & 290.32 & 88.00 & 3318.15 & 63.00 & 4600.75 & 37.42 & 5585.19 & 42.99 & 5575.36 \\
    prompt to concise~\cite{p_concise} & 94.00 & \textbf{182.42} & 90.50 & 2161.51 & 54.50 & \underline{3286.95} & 31.98 & 4120.47 & \underline{45.50} & 5611.31 \\
    L1-max~\cite{l1}          & 93.40 & 1198.63 & 87.80 & 1624.55 & 60.60 & 3832.18 & 25.81 & 3769.06 & 37.76 & 4983.77 \\
    s1-32B~\cite{s1}               & \textbf{96.00} & 1023.26 & 90.00 & 2922.14 & 53.96 & 5958.03 & \underline{60.72} & 3402.53 & 31.96 & 4603.04 \\
    DEER~\cite{deer}                 & 91.00 & 392.25 & 90.40 & \textbf{1280.75} & 51.93 & 4112.46 & 54.20 & \underline{3402.51} & 37.05 & \underline{4341.08} \\
    SEAL~\cite{seal}                 & 89.30 & 279.49 & 84.00 & 2096.87 & 42.34 & 6354.78 & 56.13 & 3468.96 & 39.03 & 5709.02 \\
    Specutive Thinking~\cite{specutive_thinking}    & 94.45 & 899.83 & 90.40 & 1943.84 & \underline{64.60} & 4941.73 & 51.37 & 5031.91 & 40.07 & 6265.42 \\
    SpecReason~\cite{specreason}           & \textbf{96.00} & 1075.98 & \underline{92.00} & 2380.78 & 43.00 & 6354.78 & 44.43 & 4885.98 & 42.63 & 5589.44 \\
    \cellcolor{pink!30}\methodname{} (Ours) & \cellcolor{pink!30}\underline{95.60} & \cellcolor{pink!30}\underline{197.03} & \cellcolor{pink!30}\textbf{93.20} & \cellcolor{pink!30}\underline{1485.36} & \cellcolor{pink!30}\textbf{67.40} & \cellcolor{pink!30}\textbf{2935.83} & \cellcolor{pink!30}\textbf{65.22} & \cellcolor{pink!30}\textbf{3024.81} & \cellcolor{pink!30}\textbf{57.11} & \cellcolor{pink!30}\textbf{4023.37} \\
    \bottomrule
  \end{tabular}
  \vspace{-5pt}
\end{table*}

\subsection{Experiment Setting}
\textbf{Datasets.}
{We consider three families of reasoning-intensive tasks: mathematical reasoning, multidisciplinary scientific question answering (QA), and code generation. For mathematical reasoning, we use \textsc{GSM8K}~\cite{gsm8k}, \textsc{MATH500}~\cite{math500}, and \textsc{AMC AIME}~\cite{numina_math_datasets}. For scientific QA, we report results on \textsc{GPQA Diamond}~\cite{gpqa_dia}. For code generation, we use the \textsc{LiveCodeBench}~\cite{livecodebench}. For each dataset, we evaluate the performance on the test set and treat the training set as the precedent set without question overlap. More details are in Appendix~\ref{sup: data}.}

\textbf{Baselines and Metrics.}

{We compare~\methodname{} with a comprehensive set of baselines. We report representative reasoning models, including DeepSeek-R1-Distilled-Qwen2.5-7B/14B/32B~\cite{DeepSeek-R1} and QwQ-32B~\cite{qwq32b}. From Related Work in Appendix~\ref{sup: related work}, we compare with representative methods for efficient reasoning. For single-model optimization, we consider both fine-tuning-based and training-free approaches. The fine-tuning-based methods include s1-32B~\cite{s1},L1~\cite{l1}. The training-free methods include DEER~\cite{deer}, SEAL~\cite{seal}. For multi-model collaboration, we include Specutive Thinking~\cite{specutive_thinking} and SpecReason~\cite{specreason}.} {We report two metrics: (i) the average reasoning length measured at the token level, denoted as \textit{tokens};  (ii) the final answer accuracy, denoted as \textit{acc}. Details about baselines and metrics are in Appendix~\ref{sup: imp_baseines} and~\ref{sup: eval_indicator}.}

{\textbf{Implementation Details.}

For \textbf{APS}, we set $k$ to 10 and $p$ to 5~\cite{icl+ft}. We use bge-large-en-v1.5~\cite{bge_embedding} to encode embeddings. For \textbf{TIE}, LoRA is applied to $W_q$ and $W_v$ to adapt to the precedent-informed instruction and the rank is 8. During the adaptation process, the train epoch is 1 and the learning rate is $3e^{-5}$. All experiments are conducted on 4$\times$NVIDIA A800 GPUs with CUDA 11.2. Main experimental results are based on the DeepSeek-R1-Distilled-32B. Details are in Appendix~\ref{sup: impl_com}.}

\subsection{Comparison Experiment}

\textbf{Balanced Accuracy-Efficiency Trade-off across Domains.}
From Table~\ref{tab: com_exp}, \methodname{} achieves the best accuracy-efficiency trade-off on AMC AIME, GPQA Diamond, and LiveCodeBench.~\methodname{} improves accuracy by 11.61\%, while reducing average reasoning length by 28.30\% over the second-best method on LiveCodeBench. This advantage arises from the reasoning experience contained in the provided precedents and the precedent-informed reasoning enhanced through test-time adaptation, which reduces redundant exploration and validation, and achieves accurate prediction.

\textbf{Adaptive Reasoning Length by Task Difficulty.}
\methodname{} scales its reasoning length with task difficulty.~\methodname{} costs 197.03 tokens on GSM8K, increases to 1485.36 tokens on MATH500, and further to 2935.83 tokens on AMC-AIME on average. This adjustment is enabled by the reasoning patterns provided from selected examples while effectively preserving the native reasoning capabilities of the base model for harder cases.

\textbf{Advantages of High-Difficulty Tasks.} As stated before,~\methodname{} attains the highest accuracy and a shortest reasoning trace on hard multi-step datasets compared with other baselines on AMC-AIME, GPQA Diamond, and LiveCodeBench. This makes it practical for challenge math solving, scientific Q\&A, and code generation, where both correctness and inference cost are critical.

\begin{table}[t]
\centering
\caption{Statistics of the occurrence of precedent reference behaviors before and after applying \textbf{TEI}. {Details of the behavior detection are in Appendix~\ref{sup: exp_imi_behavior}.}
}
\fontsize{9}{10}\selectfont 
\setlength{\tabcolsep}{11pt}
\begin{tabular}{ l c c}
\toprule
 Dataset  &\methodname{} (w/o TLM) & \methodname{} (full) \\
\midrule  
  GSM8k           & 0.00 & 1.08  \\
  MATH500          & 0.18 & 2.64  \\
   AMC AIME         & 0.11 & 5.38  \\
GPQA Diamond          & 0.57 & 5.81  \\
LiveCodeBench            & 0.04 & 8.19  \\ 
\bottomrule
\end{tabular}
\label{tab: behavior_statistics_times_only}
\vspace{-15pt}
\end{table}

\subsection{Reasoning Behavior Analysis}
\label{exp: behavior_ana}

\begin{table*}[htbp]
  \centering
  \caption{The statistical results of attention score above average in one reasoning path in token ($\mathbf{HT.}$) and segment ($\mathbf{HS.}$) level. Best results are in \textbf{bold}. More calculation details are in Appendix~\ref{sup: details of ht&hs}.}
  \label{tab: redundant_quantity}
  
  \fontsize{9}{10}\selectfont
  \setlength{\tabcolsep}{7.5pt}

  \begin{tabular}{l *{10}{c}}
    \toprule
    Method & \multicolumn{2}{c}{GSM8K} & \multicolumn{2}{c}{MATH500} & \multicolumn{2}{c}{AMC\_AIME} & \multicolumn{2}{c}{GPQA Diamond} & \multicolumn{2}{c}{LiveCodeBench} \\
    \cmidrule(lr){2-3} \cmidrule(lr){4-5} \cmidrule(lr){6-7} \cmidrule(lr){8-9} \cmidrule(lr){10-11}
    & $\mathbf{HT.}$ & $\mathbf{HS.}$ & $\mathbf{HT.}$ & $\mathbf{HS.}$ & $\mathbf{HT.}$ & $\mathbf{HS.}$ & $\mathbf{HT.}$ & $\mathbf{HS.}$ & $\mathbf{HT.}$ & $\mathbf{HS.}$ \\
    \midrule
    DeepSeek-R1-32B & 65.81 & 65.03 & 26.83 & 30.86 & 19.17 & 22.56 & 14.11 & 18.57 & 11.82 & 18.44 \\
    l1-max & 60.20 & 64.90 & 41.97 & 37.78 & 23.06 & 28.00 & 36.96 & 37.20 & 33.63 & 29.17 \\
    s1-32B & 73.97 & 69.22 & 59.92 & 39.18 & 32.15 & 33.64 & 42.46 & 40.65 & 34.79 & 35.56 \\
    DEER & 77.00 & 71.80 & 59.10 & 41.18 & 39.62 & 44.97 & 40.08 & 38.46 & 34.39 & 30.42 \\
    SEAL & 77.46 & 76.07 & 53.01 & 41.91 & 44.22 & 44.97 & 37.02 & 39.46 & 34.73 & 31.72 \\
    Speculative Thinking & 63.48 & 68.39 & 35.28 & 43.44 & 27.98 & 25.05 & 19.18 & 22.90 & 24.65 & 24.86 \\
    SpecReason & 68.16 & 64.90 & 44.36 & 53.83 & 37.46 & 35.91 & 29.44 & 35.68 & 28.53 & 32.62 \\

    \cellcolor{pink!30}\methodname{}(Ours) & \cellcolor{pink!30}\textbf{88.74} & \cellcolor{pink!30}\textbf{80.11} & \cellcolor{pink!30}\textbf{62.27} & \cellcolor{pink!30}\textbf{56.24} & \cellcolor{pink!30}\textbf{48.26} & \cellcolor{pink!30}\textbf{46.35} & \cellcolor{pink!30}\textbf{47.63} & \cellcolor{pink!30}\textbf{44.71} & \cellcolor{pink!30}\textbf{44.72} & \cellcolor{pink!30}\textbf{39.71} \\
    \bottomrule
  \end{tabular}
\end{table*}

\begin{table*}[!t]
  \centering
  \caption{Time (seconds) and GPU memory (GB) cost of different methods across datasets. Within an A800,~\methodname{} is the second-fastest method in most case, with latency close to DEER (best) and substantially faster than the base model. With accuracy improvements over DEER (Table~\ref{tab: com_exp}), \methodname{} offers a favorable accuracy--efficient trade-off.}

  \label{tab: time_gpu_metrics}
  
  \fontsize{9}{10}\selectfont

  \setlength{\tabcolsep}{5.8pt}
  
  \begin{tabular}{lcccccccccc}
    \toprule
    Method & \multicolumn{2}{c}{GSM8K} & \multicolumn{2}{c}{MATH500} & \multicolumn{2}{c}{AMC AIME} & \multicolumn{2}{c}{GPQA Diamond} & \multicolumn{2}{c}{LiveCodeBench} \\
    \cmidrule(lr){2-3} \cmidrule(lr){4-5} \cmidrule(lr){6-7} \cmidrule(lr){8-9} \cmidrule(lr){10-11}
    & \textbf{Time} & \textbf{GPU} & \textbf{Time} & \textbf{GPU} & \textbf{Time} & \textbf{GPU} & \textbf{Time} & \textbf{GPU} & \textbf{Time} & \textbf{GPU} \\
    \midrule
    DeepSeek-R1-32B & \underline{18.06} & 72.00 & 114.40 & 72.00 & 174.35 & 72.00 & 224.00 & 72.00 & 246.72 & 72.00 \\
    QwQ-32B & 59.35 & 72.00 & 145.67 & 72.00 & 211.96 & 72.00 & 253.27 & 72.00 & 253.06 & 72.00 \\
    prompt to concise & 27.43 & 72.00 & 104.15 & 72.00 & 142.28 & 72.00 & 161.92 & 72.00 & 242.33 & 72.00 \\
    s1-32B & 140.48 & 72.00 & 206.75 & 72.00 & 226.54 & 72.00 & 49.05 & 72.00 & 191.75 & 72.00 \\
    l1-max & 23.19 & 40.00 & 51.15 & 40.00 & \textbf{49.30} & 40.00 & 58.34 & 40.00 & 62.07 & 40.00 \\
    DEER & \textbf{15.23} & 72.00 & \textbf{21.75} & 72.00 & 67.73 & 72.00 & \textbf{47.26} & 72.00 & \textbf{53.90} & 72.00 \\
    SEAL & 45.61 & 72.00 & 76.43 & 72.00 & 247.40 & 72.00 & 146.21 & 72.00 & 254.14 & 72.00 \\
    Speculative Thinking & 139.23 & 112.00 & 192.15 & 112.00 & 291.64 & 112.00 & 243.27 & 112.00 & 164.68 & 112.00 \\
    SpecReason & 129.15 & 112.00 & 61.38 & 112.00 & 250.09 & 112.00 & 180.86 & 112.00 & 208.46 & 112.00 \\

    \cellcolor{pink!30}\methodname{} (Ours) & \cellcolor{pink!30}21.66 & \cellcolor{pink!30}73.18 & \cellcolor{pink!30}\underline{46.65} & \cellcolor{pink!30}74.73 & \cellcolor{pink!30}\underline{54.57} & \cellcolor{pink!30}75.62 & \cellcolor{pink!30}\underline{48.05} & \cellcolor{pink!30}76.32 & \cellcolor{pink!30}\underline{55.80} & \cellcolor{pink!30}77.15 \\
    \bottomrule
  \end{tabular}
\vspace{-10pt}
\end{table*}

\textbf{Test-Time Adaptation Promotes Precedent-Informed Reasoning.} From Table~\ref{tab: behavior_statistics_times_only}, \textbf{TEI} increases the frequency of precedent reference behavior while passive prompting with precedents triggers few such behaviors. The result indicates the effectiveness of \textbf{TEI} increases active precedent reference behaviors, which correlates with shorter reasoning trace.

{\textbf{Attention-based Evidence that Shortening Comes from Redundancy Reduction.} To verify that shorter reasoning results from removing redundancy~\cite{deer}, we compute token and segment-level salience ratios ($\mathbf{HT.}$ and $\mathbf{HS.}$) based on above-average attention to the \texttt{</think>} token (details in Appendix~\ref{sup: ht&hs})~\cite{think_clear}. From Table~\ref{tab: redundant_quantity}, the base model exhibits a long-tail pattern on hard benchmarks that both $HT.$ and $HS.$ are less than 20\%~\cite{longtail1,longtail2}. While the compared methods improve the $HT.$ and $HS.$,~\methodname{} consistently achieves best across datasets, indicating its shortens reasoning by discarding low-impact redundancy.}

{\textbf{Acceptable Time and Computational Cost when Considering Accuracy-Efficient Trade-off.} {From Table~\ref{tab: time_gpu_metrics}, the input augmentation and adaptation introduce extra time cost, which is obvious on the easy task dataset GSM8K with 18.06s by the base model and 21.66s by~\methodname{}. However, on a challenging dataset, such as AMC AIME,~\methodname{} reduces the overall latency from 174.35\,s to 54.57\,s, yielding a 3.19$\times$ speedup over the base model. Furthermore, all methods complete reasoning within an NVIDIA A800,~\methodname{} requires 73.18 to 77.15\,GB GPU memory, which is comparable to the base model (72\,GB). 

~\methodname{} is practically deployable under realistic GPU constraints and acceptable time cost when considering the outstanding reasoning performance. Stage-wise time cost of~\methodname{} is in Appendix~\ref{sup: stage_time}.}}

\subsection{Ablation Study}

\begin{table*}[t]
  \centering
  \caption{Ablation results. \emph{w/o~reference example} removes reference examples from input context. \emph{w/o instruction adaptation} disables test-time adaption and uses only precedent informed prompting for reasoning.}

  \label{tab: abs_study}
  
  \fontsize{9}{10}\selectfont
  
  \setlength{\tabcolsep}{6.4pt} 

  \begin{tabular}{lcccccccccc}
    \toprule
    Method & \multicolumn{2}{c}{GSM8K} & \multicolumn{2}{c}{MATH500} & \multicolumn{2}{c}{AMC AIME} & \multicolumn{2}{c}{GPQA Diamond} & \multicolumn{2}{c}{LiveCodeBench} \\
    \cmidrule(lr){2-3} \cmidrule(lr){4-5} \cmidrule(lr){6-7} \cmidrule(lr){8-9} \cmidrule(lr){10-11}
    & $\mathbf{acc.}$ & $\mathbf{token.}$ & $\mathbf{acc.}$ & $\mathbf{token.}$ & $\mathbf{acc.}$ & $\mathbf{token.}$ & $\mathbf{acc.}$ & $\mathbf{token.}$ & $\mathbf{acc.}$ & $\mathbf{token.}$ \\
    \midrule
    DeepSeek-R1-32B & 94.20 & 263.04 & 86.80 & 2269.37 & 59.92 & 3661.52 & 36.30 & 5028.59 & 41.02 & 5224.52 \\
    w/o~APS & 94.60 & 253.22 & 88.80 & 1819.20 & 61.80 & 3472.31 & 55.15 & 4219.80 & 45.02 & 5045.93 \\
    w/o~TEI & 95.40 & 281.37 & 90.60 & 3725.20 & 63.60 & 3933.19 & 56.88 & 5169.49 & 46.46 & 5383.76 \\

    \cellcolor{pink!30}\methodname{} (Ours) & \cellcolor{pink!30}95.60 & \cellcolor{pink!30}197.03 & \cellcolor{pink!30}93.20 & \cellcolor{pink!30}1485.36 & \cellcolor{pink!30}67.40 & \cellcolor{pink!30}2935.83 & \cellcolor{pink!30}65.22 & \cellcolor{pink!30}3024.81 & \cellcolor{pink!30}57.11 & \cellcolor{pink!30}4023.37 \\
    \bottomrule
  \end{tabular}
\vspace{-10pt}
\end{table*}

\textbf{Effect of Adaptive Precedent Selection.}
From Table~\ref{tab: abs_study}, removing precedents consistently degrades accuracy and efficiency. On AMC-AIME, accuracy drops by 5.60\% and the average reasoning length increases by 18.3\%. This indicates that high-quality, model-aware precedents contain experience and effectively narrow the solution search space, steering the model toward shorter and more reliable reasoning paths, especially for tasks that require multi-step deduction. 

\textbf{Effect of Test-time Experience Internalization.} Without test-time adaptation,~\methodname{} attains moderate accuracy but generates longer reasoning traces. On MATH500,~\methodname{} without test-time adaptation to instruction costs 2.5 times reasoning length while the accuracy is reduced by 2.6\%, indicating the model needs additional reasoning to understand the context and find the final answer to the question without the instruction adaptation. This verifies that test-time adaptation is the key to internalizing solution patterns from precedents, yielding high accuracy and concise reasoning.

\subsection{More Discussion}
\label{sec: more discussion}

\begin{figure}[t]
    \centering

    \begin{subfigure}[b]{1.0\linewidth} 
        \centering
\includegraphics[width=0.98\linewidth,keepaspectratio ]{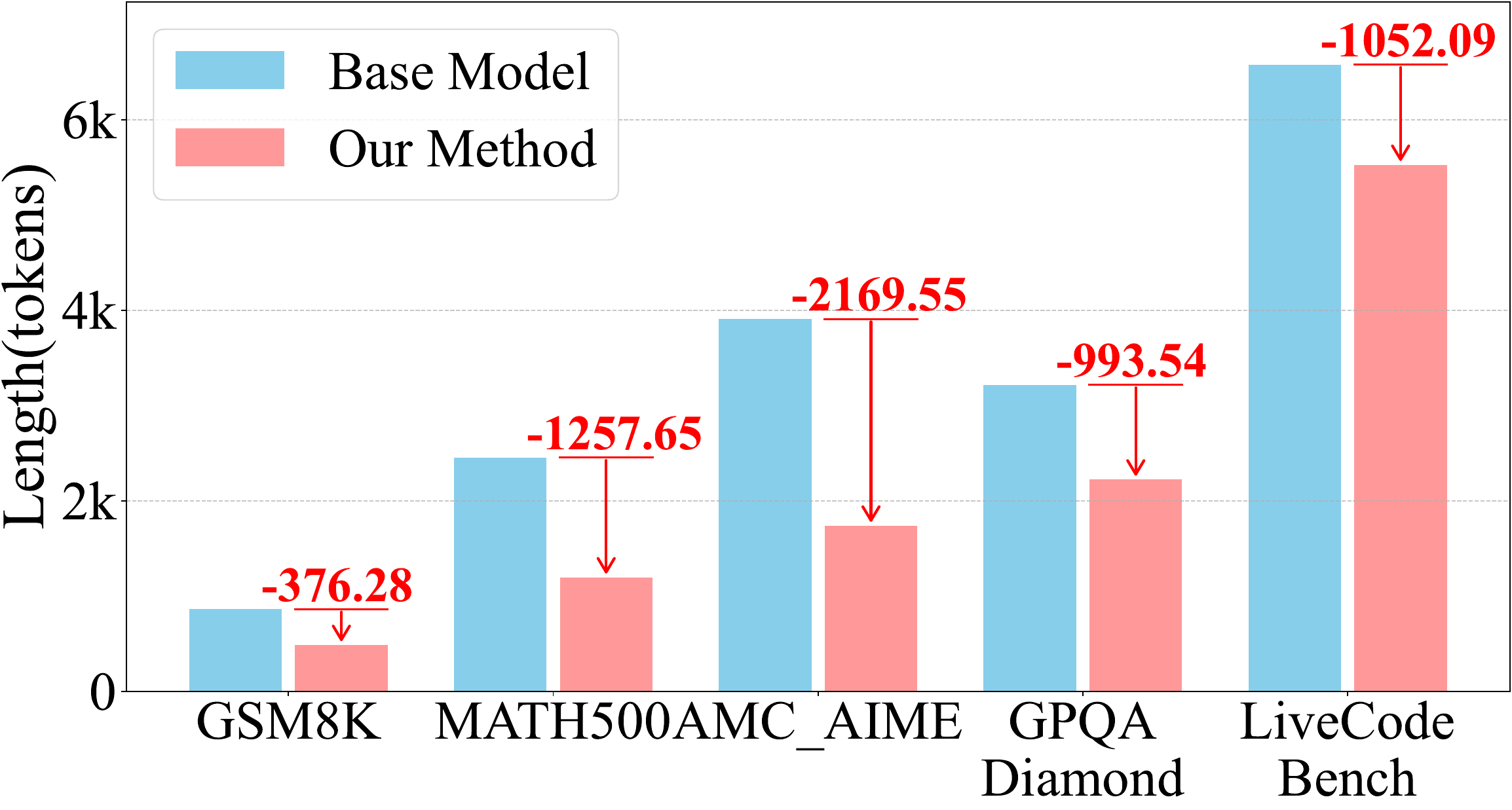} 
        \caption{The token cost for models with or without~\methodname.}
        \label{fig: token_sacl}
    \end{subfigure}

    \begin{subfigure}[b]{1.0\linewidth} 
        \centering
        \includegraphics[width=0.98\linewidth,keepaspectratio]{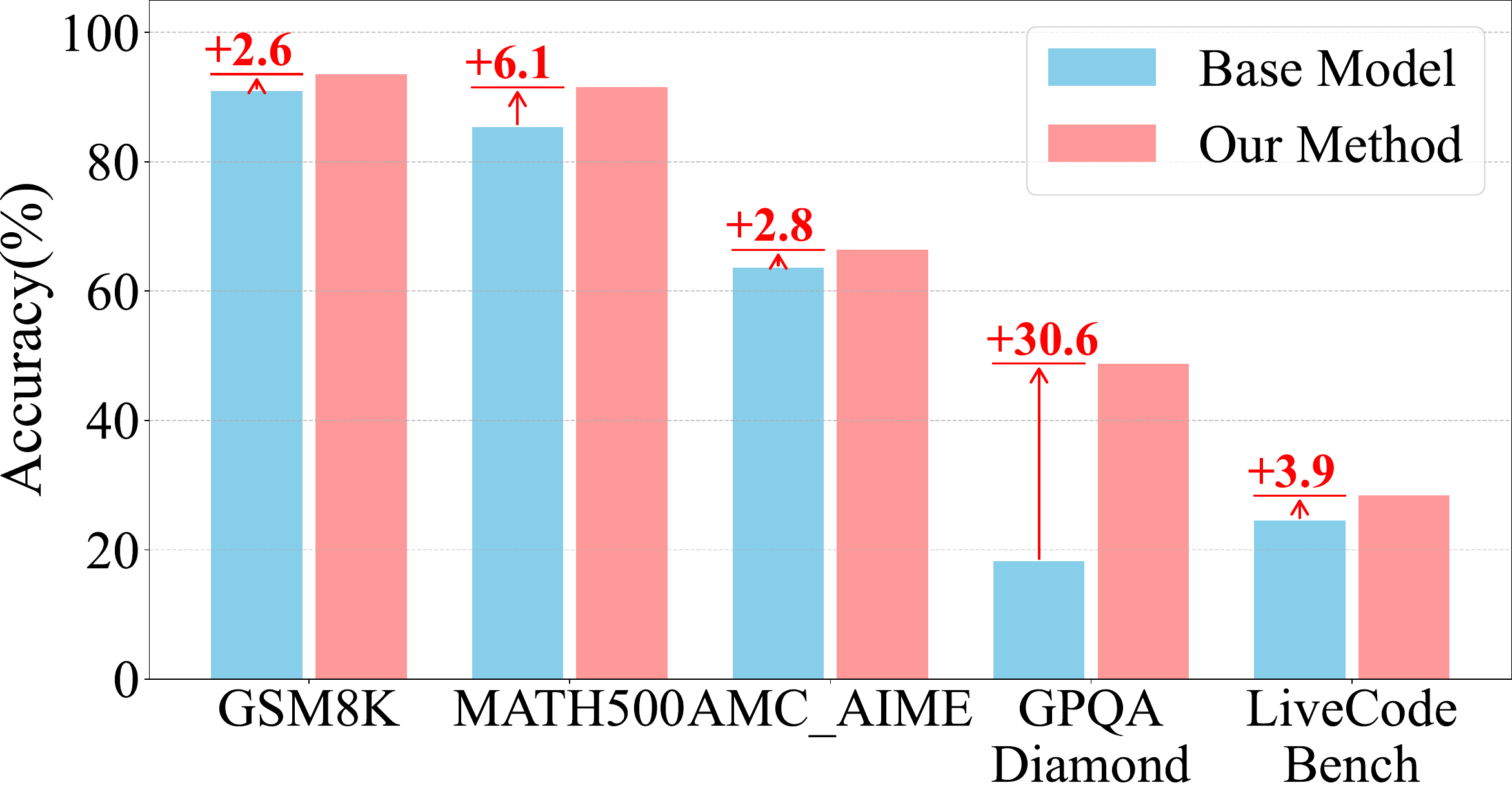} 
        \caption{The accuracy for models with or without~\methodname.}
        \label{fig: acc_sacl}
    \end{subfigure}
     \caption{Scalability of~\methodname{} across different reasoning models, with average accuracy (\%) and reasoning tokens on different models reported for each dataset.}
    \label{fig: scale}
\vspace{-15pt}
\end{figure}

\textbf{\methodname{} Scales Robustly across Reasoning Models of Different Sizes.} From 
We apply \methodname{} to QwQ-32B and DeepSeek-R1-Distilled-Qwen2.5-7B/14B and the result is shown in Figure~\ref{fig: scale}.~\methodname{} consistently reduces the reasoning length and increases accuracy across the dataset on different base models. This robustness follows from the model-aware design that the reference examples are selected by jointly considering semantic relevance and model familiarity, and test-time adaptation adapts the base model to the precedent-informed reasoning with model-specific signals. {Case study of reasoning results is in Appendix~\ref {sup: case_study}.}

\begin{table}[t]
    \centering

    \caption{Comparison of example selection strategies under the same test-time experience internalization setting. We isolate the effect of \methodname{}'s dual-dimension, adaptive selection by contrasting it with fixed-size ($e=n$) and single-dimension strategies (random, semantic-only, perplexity-only and SIFT~\cite{SIFT}). See Appendix~\ref{sup: various selection} for details of settings.}
    \label{tab: sampling_strategy_perf}
    \fontsize{8pt}{9pt}\selectfont
    \renewcommand{\tabcolsep}{1pt} 
    
    \begin{tabular*}{\linewidth}{@{\extracolsep{\fill}}l c c c c c c}
        \toprule
        Sampling & \multicolumn{2}{c}{MATH500} & \multicolumn{2}{c}{GPQA Diamond} & \multicolumn{2}{c}{LiveCodeBench} \\
        \cmidrule(lr){2-3} \cmidrule(lr){4-5} \cmidrule(lr){6-7}
         Strategy & $\mathbf{acc.}$ & $\mathbf{token.}$ & $\mathbf{acc.}$ & $\mathbf{token.}$ & $\mathbf{acc.}$ & $\mathbf{token.}$ \\
        \midrule
        
         e=0 & 88.80 & \underline{1819.20} & 36.15 & 4219.80 & 40.02 & 5224.52 \\
    e=1 & 87.00 & 2517.10 & 37.21 & \underline{3291.51} & 36.54 & 5193.00 \\
    e=2 & 89.00 & 2239.31 & \underline{49.30} & 3968.34 & 40.38 & 5840.62 \\
    e=3 & \textbf{94.00} & 2345.72 & 32.79 & 4186.22 & 38.46 & 4991.40 \\
    e=4 & 90.00 & 2154.91 & 35.88 & 4173.04 & 42.31 & 4919.39 \\
    e=5 & 86.00 & 2260.14 & 33.81 & 3615.65 & 40.38 & \underline{4853.44} \\
    random & 84.00 & 2059.22 & 27.56 & 5858.57 & 19.24 & 6887.21 \\
    sematic similarity & 85.00 & 1842.22 & 20.97 & 3803.77 & \underline{46.15} & 5168.35 \\
    perplexity & 82.35 & 2815.35 & 22.68 & 5042.83 & 26.92 & 6033.10 \\
    SIFT  & 85.19 & 3390.38 & 21.38 & 6946.68 & 28.85 & 7180.35 \\
    \cellcolor{pink!30}\methodname{}(Ours) & \cellcolor{pink!30}\underline{93.20} & \cellcolor{pink!30}\textbf{1485.36} & \cellcolor{pink!30}\textbf{57.11} & \cellcolor{pink!30}\textbf{3148.82} & \cellcolor{pink!30}\textbf{54.72} & \cellcolor{pink!30}\textbf{4834.88} \\
        \bottomrule
    \end{tabular*}
\vspace{-10pt}
\end{table}

\textbf{Adaptive Example Selection Tailored Outperforms Alternative Sampling Strategies.}
Table~\ref{tab: sampling_strategy_perf} shows that~\methodname{} achieves the best performance trade-off compared with fixed-count question-model aware filtering and other single dimension example selection strategies. On LiveCodeBench, where \methodname{} attains 54.72\% accuracy with 4834.88 tokens, compared to fixed precedents amounting to 3 and 57.11\% accuracy with 3148.82 tokens on GPQA Diamond when alternating with SIFT~\cite{SIFT}. These results confirm that the dual consideration from question and model and the adaptive selection mechanism of reference examples selection are necessary to supply informative precedents for effective precedent-informed reasoning adaptation.

\section{Conclusion}
We propose Precedent Informed Reasoning that transforms LRMs’ reasoning paradigm from exhaustive self-exploration to guided learning from precedents. First, \textbf{Adaptive Precedent Selection} constructs a compact set of precedents that are semantically related and {informative} for the model. Second, \textbf{Test-time Experience Internalization} adopts test-time learning on precedent-informed instruction, updating lightweight adapters to internalize solution patterns and use them as a prior during subsequent reasoning.

\bibliography{custom}

\newpage
\appendix

\section{Limitation}
\label{sup: limitation}
We discuss limitations of~\methodname{} to encourage further research and follow-up work:
\begin{itemize}

\item \textbf{Benefit depends on precedent quality and coverage.}

\methodname{} benefits from access to relevant precedents. when coverage is limited or domains evolve rapidly, the incremental gains from retrieval and internalization may be smaller. In the worst case, the framework can fall back to the base model's inference without harming correctness, while dynamically maintained precedent set~\cite{thetaevolve} are a promising direction to improve robustness over time.

\item \textbf{Runtime overhead.}
\methodname{} introduces retrieval and a lightweight test-time internalization step, which adds latency and computational cost. In Table~\ref{tab: time_gpu_metrics}, the overhead remains practical relative to most baselines, and motivates future systems optimizations, such as caching and batched retrieval to further reduce cost.
   
\end{itemize}

\section{The Details of Pre-experiment}
\label{sup: pre-exp}

\paragraph{Prompt.}
We adopted a prompt template for integrating examples in the pre-experiment. The template instructs the model to answer the target question by actively referencing solutions provided in the input examples, to minimize exploration of alternative solution paths, and to present the final answer enclosed in \verb|\boxed{}|. The full prompt template used in the pre-experiment is included in this appendix. 

\begin{tcolorbox}[
    colback=white,       
    colframe=black,      
    title=Prompt Template for Integrating Examples,  
    fonttitle=\bfseries, 
    boxrule=0.8pt,     
    left=5pt,            
    right=5pt,
    top=5pt,
    bottom=5pt,
        lowerbox=visible,       
      fontlower=\small     
]
Answer the given question by referencing the solution in the provided examples:\\
Given question: \{Given Question\}\\
------------------\\
Provided examples:\\
Question 1: \{Question 1\}\\
\\
Solution 1:\{Solution 1\}\\
------------------\\
Question 2: \{Question 2\}\\
\\
Solution 2:\{Solution 2\}\\
------------------\\
...\\
Since the provided examples are similar to the given question, you can actively reference the solution of the provided examples to answer the given question. You can reduce the exploration of the solution to the given question and the verification of local results. Put the final answer of the given question in \\boxed\{\}.

\tcblower  
  \textit{Note}: Since the number of selected samples varies dynamically for a given question and model, the ellipsis `$\ldots$` here indicates that the quantity is uncertain.
\end{tcolorbox}

\paragraph{Pre-experiment sample size.}
The pre-experiment used 100 problem instances as a quick validation set to probe the effect of the number of provided examples on both accuracy and token consumption. For comparability, instances were sampled from the target test set (MATH500, GPQA Diamond, LiveCodeBench) such that each sampled subset comprised the same number of examples used throughout the pre-experiment analyses.

\paragraph{Experimental settings in pre-experiment.}
The pre-experiment was run under the following settings:
\begin{itemize}
  \item \textbf{Context window (max tokens):} 32k. 
  \item \textbf{Hardware:} single NVIDIA A800 (1 × A800) for rapid iteration in the pre-experiment stage. 
  \item \textbf{Decoding temperature:} $0$ (deterministic decoding) to stabilize comparisons across different example counts.
 
\end{itemize}

\paragraph{Evaluation protocol and metrics.}
For each model and for each example-count condition \(e\in\{0,1,\dots,5\}\), we report two primary metrics: accuracy (acc) and average generated length (tokens). Tables in this appendix (Tables~6–8) present the pre-experiment results for three model configurations (DeepSeek-R1-Distilled-Qwen2.5-32B, QwQ-32B, and DeepSeek-R1-Distilled-Qwen2.5-7B) across the three task families (MATH500, GPQA Diamond, LiveCodeBench). The detailed evaluation results are in Table~\ref{tab: pre_7b}, Table~\ref{tab: pre_32b} and Table~\ref{tab: pre_qwq}.

\begin{table}[!t]
  \centering
  \caption{The result of pre-experiment for DeepSeek-R1-Distilled-Qwen2.5-7B.$e=k$ means there are k examples provided in the input prompt.}
  \label{tab: pre_7b}
  \resizebox{\linewidth}{!}{
  \begin{tabular}{l cc cc cc}
    \hline
    & \multicolumn{2}{c}{MATH500} & \multicolumn{2}{c}{GPQA Diamond} & \multicolumn{2}{c}{LiveCodeBench} \\
    & 
    $\mathbf{acc.}$ & $\mathbf{token.}$ & $\mathbf{acc.}$ & $\mathbf{token.}$ & $\mathbf{acc.}$ & $\mathbf{token.}$ \\
    \hline
    e=0 & 87.00 & 2576.49 & 41.86 & 5482.81 & 21.15 & 25058.90 \\
    e=1 & \textbf{88.00} & 2019.90 & 40.00 & 4411.38 & 19.23 & 25189.14 \\
    e=2 & 82.00 & 1754.70 & 42.00 & 4110.14 & 28.84 & \textbf{22101.56} \\
    e=3 & 82.00 & \textbf{1655.82} & 46.00 & 3943.48 & \textbf{36.53} & 23313.76 \\
    e=4 & 86.00 & 2109.38 & 48.00 & 4069.18 & 30.76 & 24710.72 \\
    e=5 & 84.00 & 1960.30 & \textbf{54.00} & \textbf{3783.68} & 32.69 & 24720.68 \\
    \hline
  \end{tabular}
  }
  
\end{table}

\begin{table}[!t]
  \centering
  \caption{The result of pre-experiment for DeepSeek-R1-Distilled-Qwen2.5-32B. $e=k$ means there are k examples provided in the input prompt.}
  \label{tab: pre_32b}
  \resizebox{\linewidth}{!}{
  \begin{tabular}{l cc cc cc}
    \hline
    & \multicolumn{2}{c}{MATH500} & \multicolumn{2}{c}{GPQA Diamond} & \multicolumn{2}{c}{LiveCodeBench} \\  &
    $\mathbf{acc.}$ & $\mathbf{len.}$ & $\mathbf{acc.}$ & $\mathbf{len.}$ & $\mathbf{acc.}$ & $\mathbf{len.}$ \\
    \hline
    e=0 & 91.00 & 3055.53 & 63.95 & 9243.30 & 55.76 & 19482.96 \\
    e=1 & 92.00 & \textbf{2285.34} & 46.51 & 8706.31 & 53.84 & 16686.50 \\
    e=2 & 88.00 & 3101.46 & 66.67 & 6267.77 & \textbf{57.69} & 14190.69 \\
    e=3 & \textbf{94.00} & 2660.18 & 55.81 & 4466.50 & 50.00 & 14973.87 \\
    e=4 & 88.00 & 3766.13 & 68.60 & 3482.04 & 48.07 & \textbf{13135.62} \\
    e=5 & 91.00 & 2732.41 & \textbf{70.93} & \textbf{3138.15} & 42.14 & 16656.68 \\
    \hline
  \end{tabular}
  }
  
\end{table}

\begin{table}[!t]
  \centering
  \caption{The result of pre-experiment for QwQ-32B. $e=k$ means there are k examples provided in the input prompt.}
   \label{tab: pre_qwq}
  \resizebox{\linewidth}{!}{
  \begin{tabular}{l cc cc cc}
    \hline
    & \multicolumn{2}{c}{MATH500} & \multicolumn{2}{c}{GPQA Diamond} & \multicolumn{2}{c}{LiveCodeBench} \\
     &
    $\mathbf{acc.}$ & $\mathbf{len.}$ & $\mathbf{acc.}$ & $\mathbf{len.}$ & $\mathbf{acc.}$ & $\mathbf{len.}$ \\
    \hline
    e=0 & 92.00 & 3671.68 & 65.11 & 9242.30 & 57.69 & 13973.98 \\
    e=1 & 91.00 & \textbf{2284.34} & 48.84 & 8705.31 & 59.61 & 14082.94 \\
    e=2 & 92.00 & 3100.47 & 66.67 & 6266.77 & \textbf{61.54} & 13747.86 \\
    e=3 & \textbf{93.00} & 2659.18 & 55.81 & 4465.50 & 50.00 & 13556.00 \\
    e=4 & 88.00 & 3765.13 & 68.60 & 3481.04 & 51.06 & \textbf{11775.93} \\
    e=5 & 84.00 & 2731.41 & \textbf{70.93} & \textbf{3137.15} & 51.92 & 11815.53 \\
    \hline
  \end{tabular}
  }
  
\end{table}

\paragraph{Brief summary of pre-experiment findings.}
Based on a rapid validation using 100 instances, we derive two practical insights: (1) the optimal number of examples relies on the interaction between model capacity and task type, implying that a fixed example count does not uniformly improve performance; (2) in specific combinations of models and tasks, adding examples can increase token consumption or even slightly reduce accuracy. These observations guided the configuration choices for larger-scales experiments reported in the main paper.

\subsection{Instruction Template in~\methodname{}}
\label{sup: ins_template}

\begin{tcolorbox}[
    colback=white,      
    colframe=black,   
    title=Prompt Template for Integrating Examples, 
    fonttitle=\bfseries, 
    boxrule=0.8pt,       
    left=5pt,         
    right=5pt,
    top=5pt,
    bottom=5pt,
        lowerbox=visible,        
      fontlower=\small]
Answer the given question by referencing the solution in the provided examples:\\
Given question: \{Given Question\}\\
------------------\\
Provided examples:\\
Question 1: \{Question 1\}\\
\\
Solution 1:\{Solution 1\}\\
------------------\\
Question 2: \{Question 2\}\\
\\
Solution 2:\{Solution 2\}\\
------------------\\
...\\
Since the provided examples are similar to the given question, you can actively reference the solution of the provided examples to answer the given question. You can reduce the exploration of the solution to the given question and the verification of local results. Put the final answer of the given question in \\boxed\{\}.

\tcblower 
  \textit{Note}: Since the number of selected samples varies dynamically for a given question and model, the ellipsis `$\ldots$` here indicates that the quantity is uncertain.
\end{tcolorbox}

\section{The Details of Dataset}
\label{sup: data}
We evaluate our methods on a set of widely used benchmark datasets. Concretely, our adopted datasets include GSM8K~\cite{gsm8k}, MATH500~\cite{math500}, AMC AIME~\cite{numina_math_datasets}, GPQA Diamond~\cite{gpqa_dia}, and LiveCodeBench~\cite{livecodebench}. AMC AIME is a combined dataset derived from NuminaMath-CoT~\cite{numina_math_datasets}, consisting of AMC and AIME problem sets collected from the \href{https://artofproblemsolving.com/wiki/}{AoPS Wiki} with the reference solutions realigned into a chain-of-thought (CoT) format using GPT-4o. For comparison and controlled evaluation, we construct precedent set and test set for each dataset with no instance overlap. The specific number of precedent and test samples for each dataset is presented in Table~\ref {tab: instance_details}. The details of the sources of precedent and test instances for each dataset are as follows:

\begin{table}[!t]
  \centering
  \caption{Sample Counts of Different Datasets for Precedent and Test Samples} 
  \label{tab: instance_details} 
    \fontsize{7pt}{8pt}\selectfont
    \setlength{\tabcolsep}{1pt} 
  \begin{tabular}{lccccc} 
    \toprule
     & GSM8K & MATH500 & AMC AIME & GPQA Diamond & LiveCodeBench \\
    \midrule 
    precedent & 2000 & 2000 & 2000 & 700 & 880 \\
    test      & 500  & 500  & 500  & 200 & 175 \\
    \bottomrule
  \end{tabular}
\end{table}

\begin{itemize}
  \item \textbf{GSM8K}\cite{gsm8k}. Grade School Math 8K, a widely adopted standard benchmark for evaluating models' fundamental arithmetic thinking capabilities, consisting of grade-school math word problems. The precedent set and the test set in paper are the corresponding training and test set from NuminaMath~\cite{numina_math_datasets} with GPT-4o to reformat reference solutions from the original problems into the CoT format.
  
  \item \textbf{MATH500}\cite{math500}. MATH is a challenging dataset for evaluating advanced mathematical reasoning, covering algebra, geometry, number theory, and calculus. It requires complex, multi-step deductions and serves as a difficult benchmark. In our experiment, we use the MATH500~\cite{math500} subset. The MATH500 subset used in our experiments can be found on its \href{https://huggingface.co/datasets/HuggingFaceH4/MATH500}{Hugging Face page}. The precedent set of this dataset is from the corresponding training set in NuminaMath~\cite{numina_math_datasets}.
  
  \item \textbf{AMC AIME}\cite{numina_math_datasets}. A subset of AMC and AIME, encompasses approximately 4,300 curated problems spanning diverse mathematical domains including arithmetic, geometry, combinatorics, and algebra. These problems, sourced from \href{https://artofproblemsolving.com/wiki/}{AoPS Wiki}, are characterized by well-defined answers in either multiple-choice (AMC) or integer (AIME) format. The difficulty ranges from intermediate (AMC 10/12 level) to highly challenging (AIME-level), requiring multi-step reasoning and non-trivial problem-solving strategies. The The precedent set and the test set in paper are the corresponding training and test set from NuminaMath~\cite{numina_math_datasets}. To ensure dataset quality and avoid overlap with existing benchmarks such as MATH, we perform embedding-based decontamination, retaining only non-overlapping instances. Solutions are reformatted into chain-of-thought style using GPT-4o.

  \item \textbf{GPQA Diamond~\cite{gpqa_dia}.} GPQA Diamond is a challenging domain-specific question collection focused on physics, chemistry, and biology reasoning problems. The questions are expert-crafted to be “Google-proof”, requiring graduate-level domain knowledge and deep reasoning, such that even well-informed non-specialists achieve low accuracy. In this paper, the test set is GPQA Diamond and the precedent set is GPQA main. GPQA main dataset is available at \href{https://github.com/idavidrein/gpqa}{this link}.

  \item \textbf{LiveCodeBench~\cite{livecodebench}.} LiveCodeBench is a programmatic reasoning benchmark divided by difficulty levels. The difficulty tiers reflect problem complexity in real competitive programming contexts: easy corresponds to introductory-level contests, medium to intermediate challenges, and hard to advanced or expert-level tasks, often requiring non‐trivial algorithms, careful implementation, or multi‐step reasoning. We use the latest LiveCodeBench, v6 as the test set and use old LiveCodeBench including v1 to v5 as the precedent set. LiveCodeBench is available at \href{https://github.com/LiveCodeBench/LiveCodeBench}{this link}.
  
\end{itemize}

\section{Implementation Details}
\label{sup: impl_com}

\subsection{Implementation Details of~\methodname}

\label{sup: imp_ours}
All training and evaluation experiments are conducted using an NVIDIA A800 GPU, with 80 GB of memory, under CUDA version 12.1. Our method is implemented using the PyTorch framework with PyTorch, version 2.5.1. The training framework for test-time LoRA finetune follows the implementation of TLM\footnote{\url{https://github.com/Fhujinwu/TLM}}.

\subsection{Implementation Details of Baselines.} 

\label{sup: imp_baseines}

\textbf{The Deployment Details of Model Baselines.} We evaluated several publicly available foundation models. For all foundation models listed below, inference was conducted with \textbf{temperature set to 0}, \textbf{top-p set to 1.0}, and utilized the \textbf{vLLM inference framework}:

\begin{itemize}
    \setlength{\leftmargin}{2em} 
    \item QwQ-32B
    \item DeepSeek-R1-Distill-Qwen-7B
    \item DeepSeek-R1-Distill-Qwen-14B
    \item DeepSeek-R1-Distill-Qwen-32B
\end{itemize}

\textbf{The Implementation Details of Prompt to think concise.}

At inference time (no additional training), we follow the concise thinking instruction~\cite{p_concise} to prompt the base model, \textbf{DeepSeek-R1-Distill-Qwen-32B} to minimize unnecessary intermediate steps and stop once the solution is determined, while preserving correctness with an 8k window size. The detailed prompt is as follows:

\begin{tcolorbox}[title=The Prompt for Thinking Concise]
\{question\} Think briefly about the answer.
\end{tcolorbox}

\textbf{The Implementation Details of L1.} L1 utilizes Length Controlled Policy Optimization (LCPO), a reinforcement learning technique, to train language models capable of adhering to specified generation length constraints during reasoning tasks. We specifically compare against the \textbf{L1-Max} variant, designed to control the maximum length of the generated output (e.g., Chain-of-Thought) while maintaining task accuracy. For our reproduction experiments using L1-Max: the maximum target token length (`ngold`) was set to \textbf{8192}, including the input prompt; inference employed deterministic decoding with \textbf{temperature set to 0}.

\textbf{The Implementation Details of Speculative Thinking.} The Speculative Thinking framework enhances a small model by delegating difficult reasoning spans to a larger model via structural cues and reflective phrases—using a dual-model setup: \textbf{DeepSeek-R1-Distill-Qwen-7B} as the speculative model ($M_S$, 1×A800) and \textbf{DeepSeek-R1-Distill-Qwen-32B} as the target ($M_T$, 1×A800). Both run on vLLM~\citep{vllm} with tensor parallelism 6, and we sample 5 completions per instance. All intervention rules follow \citep{speculative}, including trigger lexicons (affirmative/reflective/verification), takeover token budgets $(n_1,n_2,n_3)$, and the reflection cap $C_{\text{thresh}}$.

\textbf{The Implementation of s1-32B.} We apply s1-32B\footnote{\url{https://github.com/simplescaling/s1}}
with window size 8192 and temperature 0. The model runs by vLLM on NVIDIA A800*1.

\textbf{The Implementation of DEER.} We develop DEER\footnote{\url{https://github.com/yourusername/DEER}} to first install dependencies by executing pip install -r requirements.txt; it is recommended to deploy using the vLLM framework, where the path to the DeepSeek-R1-Distilled-Qwen2.5-32B as core model needs to be specified during runtime, with --max\_length 8192 set to configure the window length, --temperature 0 to control the generation temperature, and parameters such as the threshold 0.95 and thinking ratio 0.6 adjusted as required, before starting the deployment via bash ./bashes/bash-vllm-deer.sh.

\textbf{The Implementation of SEAL.} We develop SEAL\footnote{\url{https://github.com/VITA-Group/SEAL}} from first install dependencies via pip install -r requirements.txt vllm==0.5.1 transformers==4.42.3 and set up latex2sympy. Use VLLM for inference with the DeepSeek-R1-Distilled-Qwen2.5-32B as core model, configuring max\_tokens=8192 (window length) and temperature=0 in the inference script. After saving results, run evaluation using evaluation/eval.sh with appropriate PROMPT\_TYPE and model path to complete the deployment.

\textbf{The Implementation of SpecReason.}

\section{The Result on 32k Window Size}

\begin{table*}[htbp]
  \centering
  \caption{Accuracy ($\mathbf{acc.}$, \%) and average reasoning length ($\mathbf{token.}$) of models under a 32k context window.}
  
  \fontsize{9}{10}\selectfont

  \setlength{\tabcolsep}{6pt}
  
  \begin{tabular}{l *{10}{c}}
    \toprule
    Method & \multicolumn{2}{c}{GSM8K} & \multicolumn{2}{c}{MATH500} & \multicolumn{2}{c}{AMC\_AIME} & \multicolumn{2}{c}{GPQA Diamond} & \multicolumn{2}{c}{LiveCodeBench} \\
    \cmidrule(lr){2-3} \cmidrule(lr){4-5} \cmidrule(lr){6-7} \cmidrule(lr){8-9} \cmidrule(lr){10-11}
    & $\mathbf{acc.}$ & $\mathbf{token.}$ & $\mathbf{acc.}$ & $\mathbf{token.}$ & $\mathbf{acc.}$ & $\mathbf{token.}$ & $\mathbf{acc.}$ & $\mathbf{token.}$ & $\mathbf{acc.}$ & $\mathbf{token.}$ \\
    \midrule
    DeepSeek-R1-7B  & 92.60 & 440.39 & 88.91 & 4519.88 & 60.20 & 9305.71 & 27.45 & 17040.71 & 37.10 & 20331.73 \\
    DeepSeek-R1-14B & 93.40 & 465.91 & 89.80 & 3973.30 & 60.20 & 7440.76 & 47.64 & 9678.51  & 47.95 & 16099.46 \\
    DeepSeek-R1-32B & 95.00 & 464.54 & 91.00 & 4020.06 & 66.00 & 7312.36 & 53.80 & 11733.01 & 53.40 & 17384.53 \\
    QwQ-32B         & 96.20 & 1261.63& 94.29 & 3875.44 & 66.90 & 6096.18 & 53.15 & 11732.34 & 57.32 & 15163.50 \\
    \bottomrule
  \end{tabular}
  \label{tab: model_performance_32k}
\end{table*}

We report the evaluation results of locally deployed models with a 32k context window on multiple datasets, in Figure~\ref{tab: model_performance_32k}. The results are broadly consistent with those in the original paper, indicating the correctness of our local software and hardware infrastructure and supporting the reliability of our other experimental findings.

\section{The Details of Evaluation Metrics}
\label{sup: eval_indicator}

We report two primary evaluation indicators in this work: (1) \textbf{accuracy} (acc) and (2) \textbf{token consumption} (tokens). Below we describe the exact computation procedures used for both metrics.

\textbf{Accuracy (acc).}
Accuracy is measured as the proportion of examples for which the model's final answer matches the reference answer. Concretely, for each test instance we obtain the model's final answer string (the content returned by the model that is intended to be the solution). Prior to string comparison we apply lightweight normalization to both the model output and the reference:
\begin{enumerate}
  \item Trim leading and trailing whitespace.
  \item If the model returns an answer wrapped in \verb|\boxed{...}| (or other LaTeX box markers), we extract and compare the boxed content (i.e., compare the substring inside \verb|\boxed{...}|).
\end{enumerate}
After these normalization steps we perform an \emph{exact string match}. An example is counted as correct if and only if the normalized model string \texttt{==} the normalized reference string. The reported \(\text{acc}\) is the fraction of correct instances over the evaluation set. 

\textbf{Token consumption (tokens).}
Token consumption for a generated text is measured using the same tokenizer associated with the evaluated model. For a generated string we compute the encoded tokens with special tokens disabled and count the number of token ids. In code:

\begin{verbatim}

\end{verbatim}

In the tables we report the \emph{average} token count across the evaluation examples for each condition (e.g., per-example average token length). Note that by setting \verb|add_special_tokens=False| we intentionally exclude model-specific special tokens from the token count so the reported values reflect the tokenized length of the raw textual output only.

\section{The details of the detection of the precedent reference reasoning behavior in the reasoning path}
\label{sup: exp_imi_behavior}

We detect and quantify \emph{precedent reference reasoning behavior}, defined as segments of a generated chain of thought that reproduce patterns from provided examples, using a compact pattern-matching pipeline.

Briefly, we first solicit 5-10 prototype reference patterns from a large reference model and form a matching set \(\mathcal{P}=\{p_1,\dots,p_m\}\). In this paper, we match the pattern: (1) "...reference example..."; (2)  "...Reference example..."; (3)  "...From example..."; (4) "...from example..."; (5) "...in example..."; (6) "...In example..."

Given a generated CoT \(\mathcal{S}=(s_1,\dots,s_n)\), we mark a step \(s_i\) as imitative if it matches any \(p_j\in\mathcal{P}\) under normalized exact/regex or fuzzy n-gram criteria. The per-instance precedent reference count is
\[
C(\mathcal{S})=\sum_{i=1}^n\sum_{j=1}^m \mathbf{1}[s_i\ \text{matches}\ p_j],
\]
and we report normalized rates \(R_{\text{steps}}=C(\mathcal{S})/n\) and \(R_{\text{tokens}}=C(\mathcal{S})/T(\mathcal{S})\) (with \(T(\mathcal{S})\) the token length using the tokenizer in Appendix~\ref{sup: eval_indicator}).

The detection pipeline applies lightweight normalization (lowercasing, whitespace collapse, strip formatting), exact/regex checks for phrase- and structure-based patterns, and a fuzzy n-gram overlap test to capture paraphrased precedent reference. We count each match (optionally collapsing multiple matches per step) and report mean and standard deviation of \(C(\mathcal{S})\), \(R_{\text{steps}}\), and \(R_{\text{tokens}}\) across conditions. Spot checks were used to validate the detector and to tighten rules where common mathematical phrases caused false positives. This conservative, scalable procedure yields interpretable, lower-bound estimates of example-influenced reasoning.

\begin{table*}[htbp]
  \centering
  \caption{Stage time analysis. $\mathbf{Adapt.}$ and $\mathbf{Reason.}$ represent time adaptation and time reasoning respectively.}
  \label{tab: stage_time}
  
  \fontsize{9}{10}\selectfont

  \setlength{\tabcolsep}{3.5pt}
  
  \begin{tabular}{lcccccccccc}
    \toprule
    Method & \multicolumn{2}{c}{GSM8K} & \multicolumn{2}{c}{MATH500} & \multicolumn{2}{c}{AMC\_AIME} & \multicolumn{2}{c}{GPQA Diamond} & \multicolumn{2}{c}{LiveCodeBench} \\
    \cmidrule(lr){2-3} \cmidrule(lr){4-5} \cmidrule(lr){6-7} \cmidrule(lr){8-9} \cmidrule(lr){10-11}
    & \textbf{Adapt.} & \textbf{Reason.} & \textbf{Adapt.} & \textbf{Reason.} & \textbf{Adapt.} & \textbf{Reason.} & \textbf{Adapt.} & \textbf{Reason.} & \textbf{Adapt.} & \textbf{Reason.} \\
    \midrule
    Ours (DeepSeek-R1-32B) & 0.99 & 30.672 & 1.13 & 45.52 & 1.34 & 53.23 & 1.05 & 46.66 & 1.3 & 54.5 \\
    \bottomrule
  \end{tabular}
\end{table*}

\section{The details of the calculation of high token ($\mathbf{HT.}$) and high segment ($\mathbf{HS.}$)}
\label{sup: details of ht&hs}

In this subsection we give concise, implementation-ready definitions of the attention-based scores used to identify \emph{high-token} and \emph{high-segment} contributions, and we explain how these statistics are used to quantify redundancy in the reasoning trace. Our notation and core formulas follow the methodology introduced in \cite{think_clear}.

\subsection{The calculation of attention score in token and segment level}
\label{sup: details of attention score}

Let the model produce a reasoning trace of length \(T\) (tokens) that is periodically probed by inserting a short summarization prompt ending with a special end-of-thinking token \(\texttt{</think>}\) (see \cite{think_clear}). For each attention \emph{layer} \(\ell\in\{1,\dots,L\}\) and each \emph{head} \(h\in H\) we denote the attention weight from the \(\texttt{</think>}\) query to token \(t\) as
\[
\alpha^{(\ell,h)}_{\texttt{</think>} \to t}.
\]
Following the paper, the per-layer–per-head \emph{token importance score} is defined as
\begin{equation}
s^{(\ell,h)}_{t} \;=\; \alpha^{(\ell,h)}_{\texttt{</think>} \to t},
\end{equation}
which measures how much the injected summarizer token focuses on token \(t\) (higher values indicate greater contribution to the summarized thought). 

We typically aggregate these granular scores into a single token-level score by averaging across heads and layers:
\begin{equation}
s_{t} \;=\; \frac{1}{L\cdot |H|}\sum_{\ell=1}^{L}\sum_{h\in H} s^{(\ell,h)}_{t}.
\end{equation}

To obtain a \emph{segment/step} score, we first partition the reasoning trace into \(N\) contiguous reasoning steps (segments) \(r_1,\dots,r_N\), where each segment \(r\) is a set of consecutive tokens. The per-step (per-layer) aggregated score used in the paper is
\begin{equation}
c^{(\ell)}_{r} \;=\; \frac{1}{|H|\cdot |r|}\sum_{h\in H}\sum_{t\in r} s^{(\ell,h)}_{t},
\end{equation}
and we obtain a final segment score by averaging over layers:
\begin{equation}
c_{r} \;=\; \frac{1}{L}\sum_{\ell=1}^{L} c^{(\ell)}_{r}.
\end{equation}
Equations (1) to (4) above mirror the token / step scoring used to drive structure-aware pruning in \cite{think_clear}.

\subsection{The calculation of $\mathbf{HT.}$ and $\mathbf{HS.}$}
\label{sup: ht&hs}

We convert continuous importance scores into binary indicators for \emph{high-token} and \emph{high-segment} membership via simple thresholding. Let \(\tau_{\text{tok}}\) and \(\tau_{\text{seg}}\) denote token- and segment-level thresholds respectively. Two common and robust choices for these thresholds are:

\begin{itemize}
  \item \textbf{Percentile thresholding:} set \(\tau_{\text{tok}}\) to the \(p\)-th percentile of \(\{s_t\}_{t=1}^T\) (we use \(p=80\) by default), and set \(\tau_{\text{seg}}\) to the \(p\)-th percentile of \(\{c_r\}_{r=1}^N\).
  \item \textbf{Z-score thresholding:} set \(\tau_{\text{tok}}=\mu_s + \lambda\sigma_s\) and \(\tau_{\text{seg}}=\mu_c+\lambda\sigma_c\), where \(\mu_s,\sigma_s\) are the mean and standard deviation of \(\{s_t\}\), \(\mu_c,\sigma_c\) those of \(\{c_r\}\), and \(\lambda\) a small constant (e.g., \(\lambda=1\)).
\end{itemize}

Using a chosen thresholding rule, define the high-token set
\[
\mathrm{HT} \;=\; \{\, t : s_t \ge \tau_{\text{tok}} \,\},
\qquad
\mathrm{HT.} \;=\; \frac{|\mathrm{HT}|}{T},
\]
and the high-segment set
\[
\mathrm{HS} \;=\; \{\, r : c_r \ge \tau_{\text{seg}} \,\},
\qquad
\mathrm{HS.} \;=\; \frac{|\mathrm{HS}|}{N}.
\]
Thus \(\mathrm{HT.}\) is the fraction of tokens classified as high-contribution, and \(\mathrm{HS.}\) is the fraction of reasoning steps (segments) that are high-contribution under the same criterion.

\subsection{\texorpdfstring{How $\mathbf{HT.}$ and $\mathbf{HS.}$}{ measure redundancy in the reasoning process}}
\label{sup: long tail&ht}

Intuitively, reasoning traces with substantial redundancy exhibit many tokens and segments that receive little attention from the summarizer token \(\texttt{</think>}\); correspondingly, only a small subset of tokens/segments are truly informative. Therefore:

\begin{itemize}
  \item \textbf{Low \(\mathrm{HT.}\) or \(\mathrm{HS.}\) indicates high redundancy:} when the fraction of high-contribution tokens/segments is small, the trace contains a long tail of low-importance tokens that are likely redundant (this observation is empirically supported by \cite{think_clear}).
  \item \textbf{Complementary redundancy score:} for convenience one can report redundancy as the complement, e.g.
  \begin{align*}
    \mathrm{Redundancy}_{\text{tok}} &= 1 - \mathrm{HT.}, \\
    \mathrm{Redundancy}_{\text{seg}} &= 1 - \mathrm{HS.},
 \end{align*}
  so larger values indicate more redundancy.
  \item \textbf{Practical interpretation:} a low \(\mathrm{HS.}\) (few high-contribution steps) suggests that the model explored many low-value reasoning chunks (contiguous redundant segments). This motivates step-aware pruning: evict tokens from low-\(c_r\) steps first, as done in \cite{think_clear}.
\end{itemize}

\paragraph{Implementation notes.}  
(1) We compute all attention weights after forwarding the injected summarization prompt and reading the attention from the \(\texttt{</think>}\) query as in Eqn.~(1) above. (2) In practice we use the percentile threshold (80th) as default because it is robust across datasets; results are stable under modest changes to \(p\) or \(\lambda\). (3) All token counts and aggregations use the same tokenizer conventions described in Appendix~\ref{sup: eval_indicator}.

\begin{table*}[htbp]
  \centering
  \caption{The details result of scalability of~\methodname{} across different reasoning models.}
  \label{tab:model_perf_cross_col}

  \fontsize{9}{10}\selectfont
 
  \setlength{\tabcolsep}{2.5pt}

  \begin{tabular*}{\textwidth}{l @{\extracolsep{\fill}} *{10}{c}}
    \toprule
    Method & \multicolumn{2}{c}{GSM8K} & \multicolumn{2}{c}{MATH500} & \multicolumn{2}{c}{AMC\_AIME} & \multicolumn{2}{c}{GPQA Diamond} & \multicolumn{2}{c}{LiveCodeBench} \\
    \cmidrule(lr){2-3} \cmidrule(lr){4-5} \cmidrule(lr){6-7} \cmidrule(lr){8-9} \cmidrule(lr){10-11}
   
    & $\mathbf{acc.}$ & $\mathbf{token.}$ & $\mathbf{acc.}$ & $\mathbf{token.}$ & $\mathbf{acc.}$ & $\mathbf{token.}$ & $\mathbf{acc.}$ & $\mathbf{token.}$ & $\mathbf{acc.}$ & $\mathbf{token.}$ \\
    \midrule
    QwQ-32B & 92.40 & 1672.23 & 88.00 & 3318.15 & 64.45 & 5297.53 & 48.84 & 6132.83 & 30.77 & 6723.94 \\
    Ours(QwQ-32B) & 95.80 & 738.24 & 90.00 & 1557.84 & 68.77 & 2126.18 & 55.56 & 4033.46 & 34.62 & 5587.40 \\
    
    DeepSeek-R1-7B & 89.60 & 478.28 & 87.00 & 2576.49 & 62.53 & 2207.90 & 41.86 & 5482.81 & 15.38 & 7055.57 \\
    Ours(DeepSeek-R1-7B) & 92.60 & 362.50 & 88.80 & 2055.53 & 64.40 & 1475.00 & 45.21 & 4862.41 & 19.25 & 6568.69 \\

    DeepSeek-R1-14B & 90.80 & 439.25 & 86.00 & 2408.36 & 63.75 & 4223.15 & 31.40 & 5241.13 & 21.15 & 6433.59 \\
    Ours(DeepSeek-R1-14B) & 92.20 & 360.17 & 88.40 & 1842.08 & 66.00 & 1618.75 & 45.83 & 3576.72 & 25.00 & 5117.71 \\

    DeepSeek-R1-32B & 94.20 & 263.04 & 87.00 & 2021.68 & 59.92 & 3661.52 & 55.81 & 5307.51 & 30.77 & 6103.94 \\
    Ours(DeepSeek-R1-32B) & 95.60 & 197.03 & 91.00 & 1549.96 & 67.40 & 2935.83 & 68.60 & 3048.82 & 34.62 & 4834.88 \\
    \bottomrule
  \end{tabular*}
\end{table*}

\section{The details of the reasoning process in ~\methodname{} }
\label{sup: stage_time}
\subsection{The details of stage-wise time cost for~\methodname{}}

This subsection analyzes the stage-wise time cost of \methodname{} by decomposing inference into two phases: a lightweight adaptation stage, where the model performs test-time experience internalization via LoRA updates using dynamically selected precedents, and a reasoning stage, where the adapted model generates the reasoning trace and final answer.

Table \ref{tab: stage_time} shows that the adaptation stage introduces a consistently small overhead across all benchmarks (0.99–1.34s), while the reasoning stage dominates total latency and increases with task difficulty, ranging from about 30 seconds on GSM8K to over 50 seconds on AMC AIME and LiveCodeBench. Across datasets, adaptation accounts for only 2–3\% of the corresponding reasoning time, indicating that the additional computation introduced by test-time adaptation is negligible in practice. This efficiency is enabled by parameter-efficient LoRA updates with very few optimization steps.

Despite introducing an explicit adaptation phase, \methodname{} reduces the overall end-to-end inference time on complex tasks. By internalizing relevant problem-solving patterns before inference, the model avoids redundant self-exploration during reasoning, and the small adaptation overhead is more than offset by the resulting reduction in reasoning time. Overall, these results demonstrate that the adaptation mechanism in \methodname{} is computationally lightweight and enables a favorable trade-off between inference efficiency and reasoning effectiveness.

\begin{table}[!t]
  \centering
  \caption{Average number of dynamic precedents selected for different datasets.}
  \label{tab: dynamic_precedents_num}
  \fontsize{8pt}{9pt}\selectfont

  \setlength{\tabcolsep}{20pt}
  \begin{tabular}{lc}
    \toprule
    Dataset & Num. \\
    \midrule
    GSM8K & 1.19 \\
    MATH500 & 1.56 \\
    AMC\_AIME & 2.73 \\
    GPQA Diamond & 2.43 \\
    LiveCodeBench & 2.23 \\
    \bottomrule
  \end{tabular}
\end{table}

\subsection{The details dynamic example across datasets}

This subsection examines the dynamic behavior of precedent selection in \methodname{}. Table \ref{tab: dynamic_precedents_num} reports the average number of dynamic precedents selected per query by the Adaptive Precedent Selection (APS) module.

Across all benchmarks, the number of selected precedents remains small, ranging from 1.19 on GSM8K to 2.73 on AMC\_AIME, indicating that \methodname{} relies on a compact set of highly informative examples rather than a large demonstration set. More complex datasets (e.g., AMC\_AIME, GPQA Diamond, and LiveCodeBench) require more precedents on average than simpler ones, reflecting the adaptive nature of APS: harder problems benefit from multiple complementary solution patterns, whereas simpler problems are often sufficiently guided by a single precedent.

By selecting a small yet informative set of precedents and internalizing them via test-time LoRA adaptation, \methodname{} transforms external demonstrations into a short-term behavioral prior. This design reduces both adaptation cost and redundant exploration during inference, contributing to the overall efficiency and practicality of the method.

\section{The details of more discussion}
In this subsection, we report the implementation setting that we discussed in Section~\ref{sec: more discussion}.

\subsection{The details of the scalability experiment of~\methodname{}}
\label{sup: Scalability}

\paragraph{The Details for Scalability.}
Table~\ref{tab:model_perf_cross_col} reports the scalability evaluation of \methodname{} across several reasoning models and datasets. For each model we show two metrics: accuracy (acc.) and average generated token length (token.). The table demonstrates that \methodname{} consistently improves or preserves accuracy while substantially reducing token consumption across model scales. For instance, applying our method to QwQ-32B increases GSM8K accuracy from 92.4\% to 95.8\% while shrinking average token length from 1672.23 to 738.24; similarly, on DeepSeek-R1-32B the accuracy on MATH500 increases from 87\% to 91\% with a concomitant token reduction, from 2021.68 to 1549.96). These patterns hold across other datasets (AMC\_AIME, GPQA Diamond, LiveCodeBench), indicating that the approach scales favorably with model capacity and dataset type: larger gains in token efficiency are observed when the baseline trace contains more redundant exploration, and accuracy improvements are largest when the integrated examples closely match the target problem structure. The full table provides per-dataset and per-model numbers used for the quantitative comparisons in Section~\ref{sec: more discussion}.

\subsection{The Details of Comparison with Alternate Example Selection}
\label{sup: various selection}

We evaluate several alternative strategies for selecting reference examples in our experiments; below we briefly summarize the concrete implementations used for the comparisons reported in the paper.

\begin{itemize}
  \item \textbf{Fixed $e=k$ examples.} For a given target question we construct the prompt using a deterministic, fixed number \(e=k\) of examples. The selection pipeline applies the same retrieval and ranking procedure used throughout the paper and then deterministically takes the top \(k\) items to form the prompt. This strategy isolates the effect of example count while keeping the retrieval policy constant.
  \item \textbf{Random selection.} For each target question we uniformly sample \(k\) examples from the available precedent set without replacement. We repeat experiments with multiple independent random seeds and report averaged metrics; this baseline measures the sensitivity of performance to unguided example choice.
  \item \textbf{SIFT-based selection.} SIFT\cite{SIFT} selects examples by scoring candidate items according to the structural-and-feature similarity criteria described in the main text (e.g., template overlap, key-phrase matches, and coarse structural alignment). We then pick the top-\(k\) candidates under the SIFT score for prompt construction. This method targets examples that are structurally most informative for the given question.
\end{itemize}

All selection variants were evaluated under the same decoding and runtime settings reported in Appendix~\ref{sup: pre-exp}; comparison results shown in the main paper and supplementary tables quantify the trade-offs between selection sophistication, accuracy, and token-efficiency.

\newpage

\section{Case Study of~\methodname}
\label{sup: case_study}

We report representative optimization cases for each reasoning task to illustrate the practical effectiveness of our method.

\newpage

\begin{figure*}[ht]
\subsection{Example Case in MATH500}
\begin{tcolorbox}[colback=gray!5!white,colframe=black!75!black,title=\textbf{Base Model}]

\textbf{Input:}\\[4pt]
Pat is to select six cookies from a tray containing chocolate chip, oatmeal, and peanut
butter cookies. How many different assortments of six cookies can be selected?

\vspace{6pt}
\textbf{Length of reasoning:} 1715 tokens

\vspace{8pt}
\textbf{Model Reasoning:}\\[4pt]
Okay, so Pat needs to select six cookies from a tray that has chocolate chip, oatmeal, and peanut butter cookies. There are at least six of each type, so we don't have to worry about running out of any particular kind. The question is asking how many different assortments of six cookies can be selected. Since cookies of the same type aren't distinguishable, it's a problem about combinations with repetition.

Hmm, let me think. I remember that when you have to choose items with repetition allowed, the formula is something like $\binom{n + k - 1}{k}$, where $n$ is the number of types and $k$ is the number of items to choose. In this case, $n$ would be 3 because there are three types of cookies, and $k$ is 6 because Pat is selecting six cookies.

So, plugging into the formula, it should be $\binom{3 + 6 - 1}{6}$ which simplifies to $\binom{8}{6}$. Wait, let me make sure I remember the formula correctly. Yeah, the formula for combinations with repetition is $\binom{n + k - 1}{k}$, so that's correct.

Calculating $\binom{8}{6}$. I know that $\binom{n}{k}$ is equal to $\frac{n!}{k!(n-k)!}$. So, $\binom{8}{6}$ would be $\frac{8!}{6! \cdot (8-6)!}$ which is $\frac{8!}{6! \cdot 2!}$. Let me compute that.

First, $8!$ is 40320, $6!$ is 720, and $2!$ is 2. So, 40320 divided by $(720 \times 2)$ is $\frac{40320}{1440}$. Let me do that division. 40320 divided by 1440. Hmm, 1440 times 28 is 40320 because $1440 \times 10$ is 14400, $1440 \times 20$ is 28800, and $1440 \times 8$ is 11520. Adding 28800 and 11520 gives 40320. So, 28. Therefore, $\binom{8}{6}$ is 28.

Wait, but let me think again. Is this the right approach? So, the problem is about selecting six cookies where order doesn't matter and repetition is allowed because you can have multiple cookies of the same type. So, yes, combinations with repetition is the right way to go.

Alternatively, I can think of it as a stars and bars problem. The number of ways to distribute six identical cookies into three distinct types. The formula for that is also $\binom{n + k - 1}{k}$, which is the same as before. So, that gives me the same result, 28.

Let me try to visualize it. Imagine you have six cookies, and you need to separate them into three groups: chocolate chip, oatmeal, and peanut butter. To separate them, you can use two dividers. For example, if you have six stars (representing cookies) and two bars (representing dividers), the number of arrangements is the number of ways to arrange these eight symbols, where six are stars and two are bars.

The number of such arrangements is $\binom{8}{2}$ because you choose two positions out of eight to place the bars. Wait, hold on, $\binom{8}{2}$ is 28 as well. So, that's consistent with the previous calculation. So, whether I think of it as $\binom{8}{6}$ or $\binom{8}{2}$, it's the same result because of the symmetry in combinations.

Just to make sure, let me compute $\binom{8}{2}$. $\frac{8!}{2! \cdot (8-2)!} = \frac{40320}{2 \times 720} = \frac{40320}{1440} = 28$. Yep, same answer.

So, is 28 the correct number of different assortments? Let me think of a smaller case to verify. Suppose instead of six cookies, Pat is selecting two cookies from three types. Then, the number of assortments should be $\binom{3 + 2 - 1}{2} = \binom{4}{2} = 6$. Let's list them:

1. Two chocolate chip\\
2. Two oatmeal\\
3. Two peanut butter\\
4. One chocolate chip and one oatmeal\\
5. One chocolate chip and one peanut butter\\
6. One oatmeal and one peanut butter\\

\end{tcolorbox}
\end{figure*}

\begin{figure*}[ht]
\begin{tcolorbox}[colback=gray!5!white,colframe=black!75!black]

That's six, which matches the formula. So, the formula works for a smaller case. Therefore, I can be confident that for six cookies, the number is 28.

Wait, another way to think about it is using generating functions. The generating function for each type of cookie is $1 + x + x^2 + x^3 + \cdots$ since you can take any number of each type. Since there are three types, the generating function is $(1 + x + x^2 + \cdots)^3$. We need the coefficient of $x^6$ in this expansion.

But $(1 + x + x^2 + \cdots)$ is equal to $\frac{1}{1-x}$, so the generating function becomes $\left(\frac{1}{1-x}\right)^3$. The expansion of $(1-x)^{-3}$ is the sum from $n=0$ to infinity of $\binom{n + 3 - 1}{3 - 1}x^n$, which is $\sum_{n=0}^\infty \binom{n+2}{2}x^n$. Therefore, the coefficient of $x^6$ is $\binom{6 + 2}{2} = \binom{8}{2} = 28$. So, that's another way to confirm it.

Alternatively, if I didn't remember the formula, I could try to list all possible combinations, but that would be tedious for six cookies. However, for a better understanding, maybe I can think of it as variables.

Let me denote the number of chocolate chip cookies as $c$, oatmeal as $o$, and peanut butter as $p$. We have $c + o + p = 6$, where $c$, $o$, $p$ are non-negative integers. The number of solutions to this equation is the number of different assortments.

To find the number of non-negative integer solutions, the formula is $\binom{n + k - 1}{k - 1}$, where $n$ is the total number of items, and $k$ is the number of variables. In this case, $n$ is 6, and $k$ is 3, so it's $\binom{6 + 3 - 1}{3 - 1} = \binom{8}{2} = 28$. So, that's consistent as well.

Wait, so in this case, $n$ is 6 and $k$ is 3, so the formula is $\binom{6 + 3 - 1}{3 - 1} = \binom{8}{2}$. So, that's another way to get 28.

I think I've verified it multiple ways: combinations with repetition, stars and bars, generating functions, and setting up the equation with variables. All lead to the same answer, 28. So, I feel confident that 28 is the correct number of different assortments.

Just to recap, the problem is about selecting six cookies from three types with unlimited supply (since there are at least six of each). The key is that the order doesn't matter, and we can have multiple cookies of the same type. So, it's a classic combinations with repetition problem, which is solved by the formula $\binom{n + k - 1}{k}$, where $n$ is the number of types and $k$ is the number of items to choose. Plugging in $n=3$ and $k=6$, we get $\binom{8}{6} = 28$.

Therefore, the number of different assortments is 28.

**Final Answer**
The number of different assortments of six cookies that can be selected is $\boxed{28}$.

\vspace{6pt}
\textbf{Final Answer:}\\[4pt]
Pat needs to select six cookies from a tray containing chocolate chip, oatmeal, and peanut butter cookies. Since there are at least six of each type, we don't have to worry about running out of any particular kind. The problem is about combinations with repetition, where the formula is \( \binom{n + k - 1}{k} \), with \( n \) being the number of types (3) and \( k \) being the number of items to choose (6).

Using the formula, we get:
\[
\binom{3 + 6 - 1}{6} = \binom{8}{6}
\]

Calculating \( \binom{8}{6} \):
\[
\binom{8}{6} = \frac{8!}{6! \cdot (8 - 6)!} = \frac{40320}{720 \cdot 2} = \frac{40320}{1440} = 28
\]

This result can also be verified using the stars and bars method, generating functions, and setting up the equation with variables, all leading to the same answer.

Thus, the number of different assortments of six cookies that can be selected is \(\boxed{28}\).

\end{tcolorbox}
\end{figure*}

\vspace{8pt}
\begin{figure*}[ht]
\begin{tcolorbox}[colback=gray!5!white,colframe=black!75!black,title=\textbf{Our Method}]
\textbf{Input:}\\[4pt]
\texttt{<|User|>}Given question: Answer the given question by referencing the solution in provided examples:\\
Given question: Pat is to select six cookies from a tray containing only chocolate chip, oatmeal, and peanut butter cookies. There are at least six of each of these three kinds of cookies on the tray. How many different assortments of six cookies can be selected? (Note that cookies of the same type are not distinguishable.)\\
Provided examples:
Question 1:A pizza parlor offers six toppings. What is the greatest number of four-topping pizzas that can be made such that no two pizzas have the same topping combination?

Solution: Because there are 6 choices of toppings, and each pizza must have 4 of them, there are ${6 \choose 4} = \boxed{15}$ four-topping pizzas.\\
Since the provided examples are similar to the given question, you can actively reference the solution of provided examples to answer the given question.
You can reduce the exploration of the solution to the given question and the verification of local results.
Put the final answer of the given question in $\text{\texttt{\textbackslash boxed\{\}}}$.

\vspace{4pt}
\textbf{Length of reasoning:} 1762 tokens

\vspace{4pt}
\textbf{Model reasoning:}\\
Okay, so I have this problem where Pat needs to select six cookies from a tray that has chocolate chip, oatmeal, and peanut butter cookies. There are at least six of each kind, so I don't have to worry about running out of any type. The question is asking how many different assortments of six cookies Pat can select. It also mentions that cookies of the same type aren't distinguishable, which I think means that if I pick two chocolate chip cookies, they're the same as any other two chocolate chip cookies.

Hmm, this seems similar to some problems I've heard about before, maybe something to do with combinations or permutations. \hl{Wait, the example given was about pizza toppings, and they used combinations.} Let me think about that. In the pizza problem, they had six toppings and wanted to know how many four-topping pizzas they could make without repeating any combination. They used the combination formula, which is n choose k, where n is the total number of items, and k is how many you're choosing. So, in that case, it was 6 choose 4, which is 15.

But in this cookie problem, it's a bit different because instead of choosing a specific number of toppings, we're choosing six cookies, and each cookie can be one of three types. So, it's more like a problem where you have multiple categories and you want to know how many ways you can distribute a certain number of items among those categories.

I remember something about stars and bars in combinatorics. Let me recall. The stars and bars theorem is used to find the number of ways to distribute identical objects into distinct bins. In this case, the cookies are identical within each type, so the types are the bins, and the cookies are the objects. So, we have six identical cookies (stars) and three types (bars). The formula for this is (n + k - 1) choose (k - 1), where n is the number of objects and k is the number of bins.

So, plugging in the numbers, n is 6 cookies, and k is 3 types. So, the formula becomes (6 + 3 - 1) choose (3 - 1), which simplifies to (8) choose (2). Let me calculate that. 8 choose 2 is 28. So, does that mean there are 28 different assortments?

Wait, let me make sure I'm applying the formula correctly. The formula is (n + k - 1) choose (k - 1), so n is 6, k is 3. So, 6 + 3 - 1 is 8, and 3 - 1 is 2. So, yes, 8 choose 2 is 28. That seems right.

\hl{Another example: three types of cookies and choosing three cookies. The formula would be (3 + 3 - 1) choose (3 - 1) = 5 choose 2 = 10. Let me see if that makes sense. The possible combinations would be:}

3 chocolate chip\\
3 oatmeal\\
3 peanut butter\\
2 chocolate chip and 1 oatmeal

\end{tcolorbox}
\end{figure*}

\begin{figure*}[ht]
\begin{tcolorbox}[colback=gray!5!white,colframe=black!75!black]
2 chocolate chip and 1 peanut butter\\
2 oatmeal and 1 chocolate chip\\
2 oatmeal and 1 peanut butter\\
2 peanut butter and 1 chocolate chip\\
2 peanut butter and 1 oatmeal\\
1 of each type\\
That's 10 combinations, which matches the formula. So, the formula seems to hold.

Therefore, going back to the original problem, with six cookies and three types, the number of different assortments is 28. So, I think that's the answer.

Final Answer\\
The number of different assortments of six cookies that can be selected is \boxed{28}.

\vspace{4pt}
\textbf{Final Answer:} \\[4px]
To determine the number of different assortments of six cookies that can be selected from a tray containing chocolate chip, oatmeal, and peanut butter cookies, we use the stars and bars theorem. This theorem helps us find the number of ways to distribute identical objects (cookies) into distinct bins (types of cookies).

The formula for this is given by:
\[
\binom{n + k - 1}{k - 1}
\]
where \( n \) is the number of objects (cookies) and \( k \) is the number of bins (types of cookies). In this problem, \( n = 6 \) and \( k = 3 \).

Plugging in the values, we get:
\[
\binom{6 + 3 - 1}{3 - 1} = \binom{8}{2}
\]

Calculating this, we find:
\[
\binom{8}{2} = 28
\]

Thus, the number of different assortments of six cookies that can be selected is \(\boxed{28}\).
\end{tcolorbox}

\textbf{Case Analysis.}
\vspace{4pt}
\newline
\begin{minipage}[t]{0.485\textwidth} 
        This case illustrates how our method reduces unnecessary exploratory reasoning by leveraging a single relevant example.
        The baseline model generates a very long (6.8k-token) chain of thought, repeatedly revisiting derivative checks, special cases, and base-dependent interpretations.
        In contrast, our method (1) anchors the reasoning process to a structurally similar example, (2) suppresses redundant local verification, and (3) directs the model toward an appropriate algebraic pattern.
        As a result, the reasoning length decreases from \textbf{6829} to \textbf{1762} tokens (a \textbf{74\% reduction}) while preserving full correctness. This demonstrates that the proposed mechanism effectively curbs unnecessary exploration and produces concise, reliable reasoning traces.
    \end{minipage}
    \hfill 

    \begin{minipage}[t]{0.485\textwidth}

    \end{minipage}
\end{figure*}

\label{tab: temp}

\vspace{10pt}

\begin{figure*}[ht]
\subsection{Example Case in Physics of GPQA Diamond}
\begin{tcolorbox}[colback=gray!5!white,colframe=black!75!black,title=\textbf{Base Model}]
\textbf{Input:}\\[4pt]
In a parallel universe where a magnet can have an isolated North or South pole, Maxwell’s equations look different. But, specifically, which of those equations are different?

\vspace{6pt}
\textbf{Length of reasoning:} 2198 tokens

\vspace{8pt}
\textbf{Model Reasoning:}\\[4pt]
Okay, so I'm trying to figure out which of Maxwell's equations would be different in a universe where magnets can have isolated North or South poles. Hmm, interesting. I remember that in our universe, magnets always come in pairs—North and South poles. You can't have just one pole by itself. That's why when you break a magnet, you just get two smaller magnets, each with both poles. So, in this parallel universe, that rule doesn't apply. They can have isolated poles. 

Maxwell's equations are the foundation of electromagnetism, right? There are four main equations: Gauss's Law for electricity, Gauss's Law for magnetism, Faraday's Law of Induction, and Ampère's Law (with Maxwell's addition). Let me recall each of them.

1. **Gauss's Law for Electricity**: This relates the electric flux through a closed surface to the charge enclosed. The equation is $\nabla \cdot \mathbf{E} = \rho/\varepsilon_0$. This one deals with electric charges, so if there are isolated magnetic charges, does this equation change? I don't think so because this is about electric charges, not magnetic ones.

2. **Gauss's Law for Magnetism**: This states that the magnetic flux through a closed surface is zero, $\nabla \cdot \mathbf{B} = 0$. In our universe, this is because there are no magnetic monopoles—only dipoles. But in this parallel universe, they have isolated North or South poles, which are essentially magnetic monopoles. So, this equation must change. Instead of zero, it should relate to the magnetic charge density. Maybe something like $\nabla \cdot \mathbf{B} = \mu_0 \rho_m$, where $\rho_m$ is the magnetic charge density.

3. **Faraday's Law of Induction**: This describes how a changing magnetic field induces an electric field. The equation is $\nabla \times \mathbf{E} = -\partial \mathbf{B}/\partial t$. I don't think the presence of magnetic monopoles affects this directly. It's about the relationship between electric and magnetic fields in time-varying situations. So, probably this equation remains the same.

4. **Ampère's Law with Maxwell's Addition**: This relates the integrated magnetic field around a closed loop to the electric current and the rate of change of electric flux through the loop. The equation is $\nabla \times \mathbf{B} = \mu_0 \mathbf{J} + \mu_0 \varepsilon_0 \partial \mathbf{E}/\partial t$. Now, if there are magnetic charges, would this equation change? In our universe, Ampère's Law accounts for the current and displacement current. If magnetic monopoles exist, maybe there's a dual version where electric currents are related to magnetic fields. Wait, actually, in the presence of magnetic monopoles, there should be a dual version of Ampère's Law for magnetic currents. So, perhaps the equation would have an additional term involving the magnetic current density, something like $\nabla \times \mathbf{B} = \mu_0 \mathbf{J} + \mu_0 \varepsilon_0 \partial \mathbf{E}/\partial t +$ something with magnetic current. But I'm not entirely sure about the exact form. Maybe it's more about the symmetry between electric and magnetic fields.

Wait, let me think again. In our universe, Maxwell's equations are symmetric if you swap electric and magnetic fields, but only if you consider the speed of light. But since in this universe, magnetic monopoles exist, the symmetry might be more explicit. So, perhaps both Gauss's laws for electric and magnetic fields would have charge densities, and both Ampère's and Faraday's laws would have current terms. 

So, in our universe, Gauss's Law for magnetism is $\nabla \cdot \mathbf{B} = 0$ because there are no magnetic monopoles. If there are, it should be $\nabla \cdot \mathbf{B} = \mu_0 \rho_m$, similar to how Gauss's Law for electricity is $\nabla \cdot \mathbf{E} = \rho/\varepsilon_0$. 

Similarly, Ampère's Law in our universe is $\nabla \times \mathbf{B} = \mu_0 \mathbf{J} + \mu_0 \varepsilon_0 \partial \mathbf{E}/\partial t$. If there are magnetic monopoles, maybe there's a dual term. Wait, actually, in the presence of magnetic monopoles, the equations would have a dual structure. So, perhaps Ampère's Law would also include a term involving the magnetic current, similar to how Gauss's Law for magnetism includes the magnetic 

\end{tcolorbox}
\end{figure*}

\begin{figure*}[ht]
\begin{tcolorbox}[colback=gray!5!white,colframe=black!75!black]

charge density.

But I'm not sure if that's the case. Let me check my reasoning. In our universe, without magnetic monopoles, the divergence of $\mathbf{B}$ is zero. With magnetic monopoles, it's non-zero. Similarly, in our universe, the curl of $\mathbf{E}$ is related to the time derivative of $\mathbf{B}$. If there are magnetic monopoles, maybe the curl of $\mathbf{B}$ would also have a term related to the time derivative of $\mathbf{E}$, but I think that's already accounted for in Ampère's Law with Maxwell's addition. Wait, no, in Ampère's Law, the displacement current is $\partial \mathbf{E}/\partial t$, which is already there. So, maybe the change is only in Gauss's Law for magnetism.

But wait, if magnetic monopoles exist, then there should be a magnetic current, just like electric current. So, in addition to the electric current $\mathbf{J}$ in Ampère's Law, there should be a magnetic current term in Faraday's Law. Hmm, I'm getting a bit confused.

Let me try to recall. In the presence of magnetic monopoles, the equations become more symmetric. So, Gauss's Law for magnetism becomes $\nabla \cdot \mathbf{B} = \mu_0 \rho_m$, and Faraday's Law becomes $\nabla \times \mathbf{E} = -\partial \mathbf{B}/\partial t - \mu_0 \mathbf{J}_m$, where $\mathbf{J}_m$ is the magnetic current density. Similarly, Ampère's Law would have a term involving the electric current and the displacement current, but maybe also a term involving the magnetic current? Or perhaps it's the other way around.

Wait, no. Let me think about the structure. In our universe, the equations are:

1. $\nabla \cdot \mathbf{E} = \rho/\varepsilon_0$
2. $\nabla \cdot \mathbf{B} = 0$
3. $\nabla \times \mathbf{E} = -\partial \mathbf{B}/\partial t$
4. $\nabla \times \mathbf{B} = \mu_0 \mathbf{J} + \mu_0 \varepsilon_0 \partial \mathbf{E}/\partial t$

If we introduce magnetic monopoles, then equation 2 becomes $\nabla \cdot \mathbf{B} = \mu_0 \rho_m$. Then, equation 3, which is Faraday's Law, would need to account for magnetic currents. So, it would become $\nabla \times \mathbf{E} = -\partial \mathbf{B}/\partial t - \mu_0 \mathbf{J}_m$, where $\mathbf{J}_m$ is the magnetic current density. Similarly, equation 4, Ampère's Law, would need to account for electric currents and displacement current, but since we already have magnetic monopoles, maybe it remains the same? Or perhaps it also gets a term involving the magnetic current? I'm not sure.

Wait, actually, in the presence of magnetic monopoles, the equations become fully symmetric. So, both Gauss's laws have charge densities, and both curl equations have current densities and time derivatives. So, equation 3 would have a term involving the magnetic current, and equation 4 would have a term involving the electric current and displacement current. But I think the displacement current is already there, so maybe equation 4 doesn't change, but equation 3 does.

Alternatively, maybe both equations 3 and 4 get additional terms. Let me try to write them out.

With magnetic monopoles:

1. $\nabla \cdot \mathbf{E} = \rho_e / \varepsilon_0$ (unchanged)
2. $\nabla \cdot \mathbf{B} = \mu_0 \rho_m$ (changed)
3. $\nabla \times \mathbf{E} = -\partial \mathbf{B}/\partial t - \mu_0 \mathbf{J}_m$ (changed)
4. $\nabla \times \mathbf{B} = \mu_0 \mathbf{J}_e + \mu_0 \varepsilon_0 \partial \mathbf{E}/\partial t$ (unchanged, except maybe units?)

Wait, but in our universe, the displacement current is $\partial \mathbf{E}/\partial t$, which is related to the rate of change of electric field. If there are magnetic monopoles, maybe there's a similar term for the magnetic field? Or perhaps not, because the displacement current is a result of the changing electric field affecting the magnetic field, and if magnetic monopoles exist, maybe the changing magnetic field affects the electric field in a similar way, but that's already captured in Faraday's Law.

I think the main changes are in Gauss's Law for magnetism and Faraday's Law. So, equation 2 changes to include magnetic charge density, and equation 3 changes to include magnetic current density.

But I'm not entirely certain. Let me try to look for a pattern. In the absence of magnetic monopoles, the equations are asymmetric because $\nabla \cdot \mathbf{B} = 0$ and $\nabla \times \mathbf{E}$ has no current term. If we introduce magnetic monopoles, we should have $\nabla \cdot \mathbf{B} = \mu_0 \rho_m$ and $\nabla \times \mathbf{E} = -\partial \mathbf{B}/\partial t - \mu_0 \mathbf{J}_m$. So, both equations 2 and 3 change.

Wait, but in our universe, equation 4 is $\nabla \times \mathbf{B} = \mu_0 \mathbf{J} + \mu_0 \varepsilon_0 \partial \mathbf{E}/\partial t$. If we have magnetic monopoles, would equation 4 also change? Or is it only equations 2 and 3?

I think equation 4 doesn't change because it's about the magnetic field generated by electric currents and displacement currents. The presence of magnetic monopoles would affect the sources of the magnetic field (equation 2) and the induced electric fields (equation 3). So, equation 4 remains the same, but equation 3 gets an additional term for the magnetic current.

So, in summary, the two equations that would be different are:

\end{tcolorbox}
\end{figure*}

\begin{figure*}[ht]
\begin{tcolorbox}[colback=gray!5!white,colframe=black!75!black]

2. Gauss's Law for Magnetism: $\nabla \cdot \mathbf{B} = \mu_0 \rho_m$ instead of $0$.

3. Faraday's Law of Induction: $\nabla \times \mathbf{E} = -\partial \mathbf{B}/\partial t - \mu_0 \mathbf{J}_m$ instead of just $-\partial \mathbf{B}/\partial t$.

Therefore, the answer is that Gauss's Law for magnetism and Faraday's Law of Induction are different.

\vspace{6pt}
\textbf{Final Answer:}\\[4pt]
In a universe where isolated magnetic poles exist, Maxwell's equations would differ in two key aspects. The Gauss's Law for magnetism, which originally states that the divergence of the magnetic field is zero ($\nabla \cdot \mathbf{B} = 0$), would instead account for magnetic charge density, becoming $\nabla \cdot \mathbf{B} = \mu_0 \rho_m$. Additionally, Faraday's Law of Induction, which originally relates the curl of the electric field to the time derivative of the magnetic field ($\nabla \times \mathbf{E} = -\frac{\partial \mathbf{B}}{\partial t}$), would include a term for magnetic current density, resulting in $\nabla \times \mathbf{E} = -\frac{\partial \mathbf{B}}{\partial t} - \mu_0 \mathbf{J}_m$. Therefore, the two equations that are different are:\\
[4pt]
\boxed{\nabla \cdot \mathbf{B} = \mu_0 \rho_m \text{ and } \nabla \times \mathbf{E} = -\frac{\partial \mathbf{B}}{\partial t} - \mu_0 \mathbf{J}_m}
\\[4pt]
Standard Answer:
Let's call $\mathbf{E}$ and $\mathbf{B}$ the electric and magnetic fields, respectively:
The ones related to the circulation of the electric field and the divergence of the magnetic field is correct, since knowing that magnets can have an isolated pole means that magnetic monopoles exist and, thus, the contributions of magnetic charges and magnetic currents must be included in the equations. The way to include them is to "symmetry-copy" the other equations, with the following dictionary: $\mathbf{E} \leftrightarrow \mathbf{B}$; electric charge $\leftrightarrow$ magnetic charge; electric current $\leftrightarrow$ magnetic current. In this way, the equations that become modified, with added terms, are the ones related to the circulation (or curl, in differential form) of $\mathbf{E}$, and to the divergence (or flux in integral form) of $\mathbf{B}$.\\

The ones related to the divergence and the curl of the magnetic field is incorrect, because the one with the curl does not change, since it already includes all symmetric contributions appearing in its symmetric equation (curl of electric field).

The one related to the divergence of the magnetic field is incorrect because that equation does get changed, but it's not the only one; the equation for the curl (or circulation) of $\mathbf{E}$ also changes.

The one related to the circulation of the magnetic field and the flux of the electric field is incorrect because none of those equations are changed, since they already include the symmetric terms appearing in their symmetric equations (circulation of $\mathbf{E}$ and flux of $\mathbf{B}$).

The final answer is: \\[4pt]
\boxed{\parbox{1\textwidth}{\centering The ones related to the circulation of the electric field and the divergence of the magnetic field.}}

\end{tcolorbox}
\end{figure*}
\label{tab:}

\vspace{8pt}

\begin{figure*}[ht]
\begin{tcolorbox}[colback=gray!5!white,colframe=black!75!black,title=\textbf{Our Method}]
\textbf{Input:}\\[4pt]
\texttt{<|User|>}Given question: Answer the given question by referencing the solution in provided examples:
Given question: In a parallel universe where a magnet can have an isolated North or South pole, Maxwell's equations look different. But, specifically, which of those equations are different?
\\
Provided examples:
Question 1:The Lagrangian density of some system can be written as $a(F_{ij})\partial^{i}A^{j} + bA_{i}g^{ij}A_{j}$. Here, $F$ represents the electromagnetic field tensor; $\partial$ represents the partial derivative symbol, $g$ is the metric tensor, and subscripts and superscripts represent indices $i$, respectively; $a$ and $b$ are constants. What is the equation of motion of such a system?

Solution: The given lagrangian is very similar to the electromagnetic one, with one $F$ replaced by $\partial A$ (instead of $\partial A-\partial A$), and with a mass term. To solve this, one must know how to deal with indices. The equation of motion comes from applying the Euler-Lagrange equation to the lagrangian. In this equation, the term with the derivative with respect to $A$ leads to $2bA^{\alpha}$, while the other one leads to $2a(\partial^{\beta}A^{\alpha} - \partial^{\alpha}A^{\beta})$. Putting everything together, we get the answer.
The final answer is: \\[4pt]
$\boxed{A^{i} = \frac{a}{b}\partial_{j}F^{ji}}$.
\\[4pt]
Question 2:If Maxwell's equations are written in potential form, one of them is given by
$\nabla^2\mathbf{A} - \frac{\partial^2}{\partial t^2}\mathbf{A} = \nabla(\nabla\cdot\mathbf{A} + \frac{\partial}{\partial t}V) - \mathbf{J}$
(ignoring constants), where $\mathbf{A}$ and $V$ are the vector and scalar potentials, respectively, and $\mathbf{J}$ is the electric current density. Now imagine that magnetic monopoles exist, but they can only form currents (i.e., charge density of magnetic monopoles, $\rho_M = 0$, but current density of magnetic monopoles, $\mathbf{J}_M$, could be not 0). And, further, assume that the magnetic current density can be described by the curl of some vector function $\mathbf{j}_m$.
In that case, how would the previous equation change?

Solution: Since the magnetic current density, $\mathbf{J}_M$, can be written as $\nabla \times \mathbf{j}_m$, the relations between $\mathbf{E}$, $\mathbf{B}$ and the potentials become: $\mathbf{B} = \nabla \times \mathbf{A}$, $\mathbf{E} = -\nabla V -\frac{\partial}{\partial t}\mathbf{A} + \mathbf{j}_m$. Plugging this into the Maxwell equation related to the curl of $\mathbf{B}$ (the one that leads to the given equation) and doing some algebra, one gets to the addition of the $-\frac{\partial}{\partial t}\mathbf{j}_m$ term. \\
The final answer is:\\[4pt]
\boxed{\parbox{1\textwidth}{\centering A term $-\frac{\partial}{\partial t}\mathbf{j}_m$ is added to the right side of the equation.}}
\\[4pt]
Question 3:In a parallel universe where a magnet can have an isolated North or South pole, Maxwell's equations look different. But, specifically, which of those equations are different?

Solution: Let's call $\mathbf{E}$ and $\mathbf{B}$ the electric and magnetic fields, respectively:\\
The ones related to the circulation of the electric field and the divergence of the magnetic field is correct, since knowing that magnets can have an isolated pole means that magnetic monopoles exist and, thus, the contributions of magnetic charges and magnetic currents must be included in the equations. The way to include them is to "symmetry-copy" the other equations, with the following dictionary: $\mathbf{E} \leftrightarrow \mathbf{B}$; electric charge $\leftrightarrow$ magnetic charge; electric current $\leftrightarrow$ magnetic current. In this way, the equations that become modified, with added terms, are the ones related to the circulation (or curl, in differential form) of $\mathbf{E}$, and to the divergence (or flux in integral form) of $\mathbf{B}$. 
The ones related to the divergence and the curl of the magnetic field is incorrect, because the one with the curl does not change, since it already includes all symmetric contributions appearing in its symmetric equation (curl of electric field). \\
The one related to the divergence of the magnetic field is incorrect because that equation does get changed, but it's not the only one; the equation for the curl (or circulation) of $\mathbf{E}$ also changes.\\
The one related to the circulation of the magnetic field and the flux of the electric field is incorrect because none of those equations are changed, since they already include the symmetric terms appearing in their symmetric equations (circulation of $\mathbf{E}$ and flux of $\mathbf{B}$).\\

\end{tcolorbox}
\end{figure*}

\begin{figure*}[ht]
\begin{tcolorbox}[colback=gray!5!white,colframe=black!75!black]

The final answer is:\\[4pt]
$\boxed{\parbox{1\textwidth}{\centering The ones related to the circulation of the electric field and the divergence of the magnetic field.}}$.\\[4pt]
\\
Since the provided examples are similar to the given question, you can actively reference the solution of provided examples to answer the given question.
You can reduce the exploration of the solution to the given question and the verification of local results.
Put the final answer of the given question in \boxed\{\}.

Solution: Because there are 6 choices of toppings, and each pizza must have 4 of them, there are ${6 \choose 4} = \boxed{15}$ four-topping pizzas.\\
Since the provided examples are similar to the given question, you can actively reference the solution of provided examples to answer the given question.
You can reduce the exploration of the solution to the given question and the verification of local results.
Put the final answer of the given question in \boxed\{\}.

\vspace{4pt}
\textbf{Length of reasoning:} 292 tokens

\vspace{4pt}
\textbf{Model reasoning:}\\
Okay, so I have this question about Maxwell's equations in a universe where magnets can have isolated North or South poles. That means magnetic monopoles exist, right? I remember from my physics classes that Maxwell's equations assume there are no magnetic monopoles, so if they do exist, some of the equations must change.

\hl{Looking at the provided examples,} especially Question 3, the solution talks about how the existence of magnetic monopoles affects Maxwell's equations. It mentions that the equations related to the circulation of the electric field (curl E) and the divergence of the magnetic field (div B) are the ones that change. 

In the standard Maxwell's equations, the divergence of B is zero because there are no magnetic monopoles. But if monopoles exist, div B would be proportional to the magnetic charge density. Similarly, the curl of E is usually -dB/dt, but with magnetic currents, there would be an additional term related to the magnetic current density.

So, applying this to the question, the equations that would be different are the ones involving the curl of E and the divergence of B. The other equations, like the curl of B and the divergence of E, remain the same because they already account for the symmetric terms when considering electric charges and currents.

Therefore, the answer should be that the equations related to the circulation of the electric field and the divergence of the magnetic field are the ones that change.

\vspace{4pt}
\textbf{Final Answer:} \\[4px]
The equations that change are those involving the curl of the electric field and the divergence of the magnetic field, as magnetic monopoles introduce new terms into these equations.

$\boxed{\parbox{1\textwidth}{\centering The ones related to the circulation of the electric field and the divergence of the magnetic field.}}$

\end{tcolorbox}

\textbf{Case Analysis.} 
\vspace{3px}
\newline
\begin{minipage}[t]{0.485\textwidth} 
        This case illustrates how our example-guided method turns broad exploratory reasoning into a focused, correct solution with much lower cost. The baseline model executed a long, iterative chain that revisited alternative hypotheses and performed redundant checks, arriving at the right answer but with substantial wasted reasoning effort.

        In contrast, our method directly leverages structurally similar examples to constrain the search space and apply the appropriate domain-level transformation, producing a concise, correct 
    \end{minipage}
    \hfill
    \begin{minipage}[t]{0.485\textwidth}
        derivation with far fewer tokens and far less detour.

        Overall, the comparison highlights three practical benefits of the approach: (1) it enforces relevant abstractions from examples so the model reasons at the right level, (2) it substantially reduces unnecessary hypothesis testing and token use, and (3) it yields clearer, more reproducible answers — desirable properties for scientific QA and other high-precision tasks.
    \end{minipage}
\end{figure*}
\label{tab:}

\begin{figure*}[ht]
\subsection{Example Case in Medium of Livecode Bench}
\begin{tcolorbox}[colback=gray!5!white,colframe=black!75!black,title=\textbf{Base Model}]
\textbf{Input:}\\[4pt]
You are an expert Python programmer. You will be given a question (problem specification) and will generate a correct Python program that matches the specification and passes all tests.

\#\#\# Question:
You are given an n x n square matrix of integers grid. Return the matrix such that:

The diagonals in the bottom-left triangle (including the middle diagonal) are sorted in non-increasing order.
The diagonals in the top-right triangle are sorted in non-decreasing order.

Example 1:

Input: grid = [[1,7,3],[9,8,2],[4,5,6]]\\
Output: [[8,2,3],[9,6,7],[4,5,1]]\\
Explanation:

The diagonals with a black arrow (bottom-left triangle) should be sorted in non-increasing order:

[1, 8, 6] becomes [8, 6, 1].
[9, 5] and [4] remain unchanged.

The diagonals with a blue arrow (top-right triangle) should be sorted in non-decreasing order:

[7, 2] becomes [2, 7].
[3] remains unchanged.

Example 2:

Input: grid = [[0,1],[1,2]]\\
Output: [[2,1],[1,0]]\\
Explanation:

The diagonals with a black arrow must be non-increasing, so [0, 2] is changed to [2, 0]. The other diagonals are already in the correct order.

Example 3:

Input: grid = [[1]]\\
Output: [[1]]\\
Explanation:
Diagonals with exactly one element are already in order, so no changes are needed.

Constraints:
\begin{verbatim} 
grid.length == grid[i].length == n
1 <= n <= 10
-10^5 <= grid[i][j] <= 10^5
\end{verbatim}

\#\#\# Format: You will use the following 
starter code to write the solution to 
the problem and enclose your code within 
delimiters.

\begin{verbatim} 
```python
class Solution:
    def sortMatrix(self, grid: List
    [List[int]]) -> List[List[int]]:     
```
\end{verbatim}
\#\#\# Answer: (use the provided format with backticks)

\vspace{6pt}
\textbf{Length of reasoning:} 4285 tokens

\vspace{8pt}
\textbf{Model Reasoning:}\\[4pt]
Okay, I need to solve this problem where I have to sort the diagonals of a square matrix in a specific way. Let's see the problem statement again.

So, the matrix is \(n \times n\). The task is to sort the diagonals in two different ways depending on which triangle they're in. The bottom-left triangle, including the middle diagonal, should have their diagonals sorted in non-increasing order. The top-right triangle's diagonals should be sorted in non-decreasing order.

Hmm, let's break this down. First, I need to identify which diagonals belong to which triangle. Then, for each diagonal, I sort it accordingly and place it back into the matrix.

\end{tcolorbox}
\end{figure*}

\begin{figure*}[ht]
\begin{tcolorbox}[colback=gray!5!white,colframe=black!75!black]

Wait, how are the diagonals structured? For a square matrix, each diagonal can be identified by the sum of the row and column indices. For example, in a \(3\times 3\) matrix, the main diagonal has \(i + j = 2\) (since indices start at 0). The diagonals above it have lower sums, and those below have higher sums.

Wait, no. Wait, for a \(3\times 3\) matrix, the main diagonal is \(i - j = 0\). But perhaps it's easier to think in terms of the sum \(i + j\). Let me think: for a \(3\times 3\) matrix, the main diagonal is \(i + j = 2\) (since \(0+2=2\), \(1+1=2\), \(2+0=2\)). The diagonals below it (like the bottom-left triangle) would have \(i + j \textgreater= 2\). Wait, no. Wait, for a \(3\times 3\) matrix, the bottom-left triangle includes the main diagonal and the ones below it. So, for each diagonal, if \(i + j\) is greater than or equal to a certain value, it's in the bottom-left.

Wait, maybe I should think about the diagonals as starting from the bottom-left corner. So, the main diagonal is part of the bottom-left triangle. So, for each diagonal, if it's in the bottom-left triangle, we sort it in non-increasing order. Otherwise, in the top-right, we sort in non-decreasing.

So, how to determine which diagonals are in which triangle. Let's think about the sum \(i + j\). For a given diagonal, the sum is \(s = i + j\). For the bottom-left triangle, including the main diagonal, the sum \(s\) is \textgreater= (n-1) - (i - j) ? Wait, maybe I'm complicating it.

Alternatively, perhaps the bottom-left triangle consists of all cells where \(i \textgreater= j\). Because for a square matrix, the main diagonal is where \(i = j\). So, the bottom-left triangle includes all cells where \(i \textgreater= j\), and the top-right includes where \(i \textless j\).

Wait, no. Because in the first example, the grid is \(3\times 3\). The bottom-left triangle includes the main diagonal and the cells below it. So, for each cell \((i,j)\), if \(i \textgreater= j\), it's in the bottom-left. So, the diagonals for the bottom-left are those where \(i - j\) is the same. Wait, no, because the diagonals are defined by \(i - j\) being constant. So, for each diagonal, the difference \(i - j\) is the same. So, for the bottom-left triangle, the diagonals are those where \(i - j \textgreater= 0\). Because for \(i \textgreater= j\), the difference is non-negative.

Wait, no. Because for a \(3\times 3\) matrix, the main diagonal is \(i - j = 0\). The diagonals below it have \(i - j = 1, 2,\) etc. So, the bottom-left triangle includes all diagonals where \(i - j \textgreater= 0\). The top-right includes those where \(i - j \textless 0\).

Wait, but in the first example, the bottom-left triangle includes the main diagonal and the ones below. So, for each diagonal, if \(i - j\) is \textgreater= 0, it's in the bottom-left. So, for each such diagonal, we need to sort it in non-increasing order.

Wait, but how to collect the elements for each diagonal. For a given diagonal, the elements are those where \(i - j\) is a constant. So, for each possible difference \(d\), collect all elements where \(i - j = d\).

But wait, for a square matrix, the possible differences \(d\) range from \(-(n-1)\) to \(n-1\). For example, in a \(3\times 3\) matrix, \(d\) can be \(-2, -1, 0, 1, 2\).

So, for each \(d\), collect all elements where \(i - j = d\). Then, for each such diagonal, determine whether it's in the bottom-left or top-right.

Wait, but for \(d \textgreater= 0\), it's in the bottom-left. For \(d \textless 0\), it's in the top-right.

So, for each \(d\) in the range from \(-(n-1)\) to \(n-1\):

- If \(d \textgreater= 0\): collect the elements, sort in non-increasing order.
- If \(d \textless 0\): collect the elements, sort in non-decreasing order.

Wait, but wait. Let's see the first example:

Input grid is:

\begin{verbatim}
1 7 3
9 8 2
4 5 6
\end{verbatim}

The main diagonal is \(1,8,6\). So, \(d=0\). So, we sort this in non-increasing order: \(8,6,1\).

Then, the diagonal with \(d=1\): elements are \(9,5\). So, sorted in non-increasing order: \(9,5\).

Diagonal \(d=2\): \(4\). So, remains \(4\).

\end{tcolorbox}
\end{figure*}

\begin{figure*}[ht]
\begin{tcolorbox}[colback=gray!5!white,colframe=black!75!black]

For \(d=-1\): elements are \(7,2\). So, sorted in non-decreasing order: \(2,7\).

\(d=-2\): \(3\). Remains \(3\).

So, the output is:

\begin{verbatim}
8 2 3
9 6 7
4 5 1
\end{verbatim}

Which matches the example.

So, the approach is correct.

So, the steps are:

1. For each diagonal in the bottom-left triangle (\(i - j \textgreater= 0\)), collect the elements, sort them in non-increasing order, and place them back.

2. For each diagonal in the top-right triangle (\(i - j \textless 0\)), collect the elements, sort them in non-decreasing order, and place them back.

Wait, but how to collect the elements for each \(d\).

Let's think about how to collect the elements for a given \(d\).

For \(d \textgreater= 0\):

The diagonal starts at \((d, 0)\) and goes up to \((n-1, n-1 - d)\). Wait, no. Wait, for \(d=0\), the diagonal is \((0,0), (1,1), (2,2)\). For \(d=1\), it's \((1,0), (2,1)\). For \(d=2\), it's \((2,0)\).

Wait, for \(d \textgreater= 0\), the starting point is \((d, 0)\), and each step increases \(i\) by \(1\) and \(j\) by \(1\) until \(i\) reaches \(n-1\).

Wait, no. For \(d=1\), the elements are \((1,0), (2,1)\). So, for \(d=1\), the starting \(i\) is \(d\), \(j\) is \(0\). Then, \(i\) increases by \(1\), \(j\) increases by \(1\) until \(i < n\) and \(j < n\).

Wait, but for \(d=2\) in a \(3\times 3\) matrix, the starting \(i\) is \(2\), \(j\) is \(0\). Then, \(i+1\) is \(3\) which is beyond \(n-1 (2)\), so only one element.

So, for \(d \textgreater= 0\), the elements are \((d, 0), (d+1, 1), \ldots\) until \(i < n\) and \(j < n\).

For \(d \textless 0\):

The diagonal starts at \((0, -d)\) and goes down to \((n-1 + d, n-1)\). For example, \(d=-1\): starts at \((0,1)\), then \((1,2)\). For \(d=-2\): starts at \((0,2)\).

So, for \(d \textless 0\), the starting point is \((0, -d)\), and each step increases \(i\) and \(j\) by \(1\) until \(i < n\) and \(j < n\).

So, the plan is:

Loop over each possible \(d\) from \(-(n-1)\) to \(n-1\).

For each \(d\):

   collect the elements along the diagonal \(i - j = d\).

   sort the elements as per the rule.

   then, place them back into the matrix.

So, how to implement this in Python.

First, get the size \(n\) of the grid.

\begin{verbatim}
n = len(grid)
\end{verbatim}

Then, for each \(d\) in range from \(-(n-1)\) to \(n-1\):

   collect the elements.

   sort them.

   replace them.

But how to collect the elements for each \(d\).

Let's think about \(d \textgreater= 0\):

\begin{verbatim}
start_i = d

start_j = 0
\end{verbatim}

then, for each step, \(i\) increases by \(1\), \(j\) increases by \(1\), until \(i < n\) and \(j < n\).

\end{tcolorbox}
\end{figure*}

\begin{figure*}[ht]
\begin{tcolorbox}[colback=gray!5!white,colframe=black!75!black]

So, for \(d \textgreater= 0\):

\begin{verbatim}
elements = []
i = d
j = 0
while i < n and j < n:
    elements.append(grid[i][j])
    i += 1
    j += 1
\end{verbatim}

Similarly, for \(d \textless 0\):

\begin{verbatim}
start_i = 0

start_j = -d 
\end{verbatim}

then, \(i\) increases by \(1\), \(j\) increases by \(1\) until \(i < n\) and \(j < n\).

So, for \(d \textless 0\):

\begin{verbatim}
elements = []
i = 0
j = -d
while i < n and j < n:
    elements.append(grid[i][j])
    i += 1
    j += 1
\end{verbatim}

Once the elements are collected, we sort them.

If \(d \textgreater= 0\): sort in non-increasing order (reverse=True).

Else: sort in non-decreasing order (reverse=False).

Then, we need to place the sorted elements back into the grid.

So, for \(d \textgreater= 0\):

\begin{verbatim}
i = d
j = 0
k = 0
while i < n and j < n:
    grid[i][j] = sorted_elements[k]
    i += 1
    j += 1
    k += 1
\end{verbatim}

Similarly for \(d \textless 0\).

So, putting it all together.

But wait, in the code, for \(d\) in range(\(-(n-1), n\)):

Wait, in Python, the range is exclusive of the end. So, for \(n=3\), the range is from \(-2\) to \(3\), but the end is \(3\), so it's up to \(2\). So, the code would loop \(d\) from \(-2, -1, 0, 1, 2\).

Yes.

So, the code should work.

Now, let's write the code.

\begin{verbatim}
class Solution:
    def sortMatrix(self, grid: List[List[int]]) -> List[List[int]]:
        
        n = len(grid)

        for d in range(-(n-1), n):
\end{verbatim}
\end{tcolorbox}
\end{figure*}

\begin{figure*}[ht]
\begin{tcolorbox}[colback=gray!5!white,colframe=black!75!black]
\begin{verbatim}
            if d >= 0:
                i, j = d, 0
            else:
                i, j = 0, -d

            elements = []
            while i < n and j < n:
                elements.append(grid[i][j])
                i += 1
                j += 1

            if d >= 0:
                elements.sort(reverse=True)
            else:
                elements.sort()

            if d >= 0:
                i, j = d, 0
            else:
                i, j = 0, -d

            k = 0
            while i < n and j < n:
                grid[i][j] = elements[k]
                i += 1
                j += 1
                k += 1

        return grid
\end{verbatim}

Wait, but in the code, the grid is modified in place. So, the function returns the same grid object, but modified.

But in Python, when you pass a list to a function and modify it, the original list is modified. So, the function can return grid.

Testing the code with the first example.

Another test case: \(\texttt{grid = [[0,1],[1,2]]}\)

\(n=2\).

\(d\) can be \(-1, 0, 1\).

\(d=1\): elements are \((1,0)\) which is \(1\). So, sorted in non-increasing order is \([1]\).

\(d=0\): elements are \((0,0)=0, (1,1)=2\). So, sorted in non-increasing order is \([2,0]\).

\(d=-1\): elements are \((0,1)=1\). So, sorted in non-decreasing is \([1]\).

So, the grid becomes:

\begin{verbatim}
2 1
1 0
\end{verbatim}

Which is correct.

So, the code should handle all cases.

Now, let's think about the code.

But wait, in the code, the for loop is for \(d\) in range(\(-(n-1), n\)). But in Python, the range function can't handle negative starts if the end is positive. Wait, no, it can. For example, range(-2,3) gives -2, -1, 0, 1, 2.

Yes.

\end{tcolorbox}
\end{figure*}

\begin{figure*}[ht]
\begin{tcolorbox}[colback=gray!5!white,colframe=black!75!black]

So, the code is correct.

So, the final code is as written.

\vspace{6pt}
\textbf{Final Answer:}\\[4pt]
To solve this problem, we need to sort the diagonals of a given square matrix in specific ways. The diagonals in the bottom-left triangle (including the main diagonal) should be sorted in non-increasing order, while the diagonals in the top-right triangle should be sorted in non-decreasing order.

\#\#\# Approach\\
1. **Identify Diagonals**: Each diagonal in the matrix can be identified by the difference between the row and column indices (i - j). For a given diagonal, this difference is constant.\\
2. **Classify Diagonals**: Diagonals where the difference (i - j) is non-negative belong to the bottom-left triangle and should be sorted in non-increasing order. Diagonals where the difference is negative belong to the top-right triangle and should be sorted in non-decreasing order.\\
3. **Collect and Sort Elements**: For each diagonal, collect the elements, sort them according to the specified order, and then place them back into the matrix.

\#\#\# Solution Code
\begin{verbatim}
```python
from typing import List

class Solution:
    def sortMatrix(self, grid: List[List[int]]) -> List[List[int]]:
        n = len(grid)
        for d in range(-(n-1), n):
            if d >= 0:
                i, j = d, 0
            else:
                i, j = 0, -d
            elements = []
            while i < n and j < n:
                elements.append(grid[i][j])
                i += 1
                j += 1
            if d >= 0:
                elements.sort(reverse=True)
            else:
                elements.sort()
            if d >= 0:
                i, j = d, 0
            else:
                i, j = 0, -d
            k = 0
            while i < n and j < n and k < len(elements):
                grid[i][j] = elements[k]
                i += 1
                j += 1
                k += 1
        return grid
```
\end{verbatim}
\#\#\# Explanation\\

\end{tcolorbox}
\end{figure*}

\begin{figure*}[t!]
\vspace{-5cm}
\begin{tcolorbox}[colback=gray!5!white,colframe=black!75!black]

1. **Loop Through Diagonals**: The loop iterates over all possible diagonal differences from -(n-1) to n-1.\\
2. **Collect Elements**: For each diagonal, elements are collected by moving along the diagonal from the starting point until the boundaries of the matrix are reached.\\
3. **Sort Elements**: Depending on whether the diagonal is in the bottom-left or top-right triangle, the elements are sorted in non-increasing or non-decreasing order, respectively.\\
4. **Replace Elements**: The sorted elements are then placed back into their respective positions in the matrix.\\

\end{tcolorbox}
\end{figure*}
\label{tab:}

\vspace{8pt}

\begin{figure*}[t!]
\begin{tcolorbox}[colback=gray!5!white,colframe=black!75!black,title=\textbf{Our Method}]
\textbf{Input:}\\[4pt]
\texttt{<|User|>}Given question: Given question: You are an expert Python programmer. You will be given a question (problem specification) and will generate a correct Python program that matches the specification and passes all tests.

\#\#\# Question:
You are given an n x n square matrix of integers grid. Return the matrix such that:

The diagonals in the bottom-left triangle (including the middle diagonal) are sorted in non-increasing order.
The diagonals in the top-right triangle are sorted in non-decreasing order.

Example 1:

Input: grid = [[1,7,3],[9,8,2],[4,5,6]]\\
Output: [[8,2,3],[9,6,7],[4,5,1]]\\
Explanation:

The diagonals with a black arrow (bottom-left triangle) should be sorted in non-increasing order:

[1, 8, 6] becomes [8, 6, 1].
[9, 5] and [4] remain unchanged.

The diagonals with a blue arrow (top-right triangle) should be sorted in non-decreasing order:

[7, 2] becomes [2, 7].
[3] remains unchanged.

Example 2:

Input: grid = [[0,1],[1,2]]\\
Output: [[2,1],[1,0]]\\
Explanation:
The diagonals with a black arrow must be non-increasing, so [0, 2] is changed to [2, 0]. The other diagonals are already in the correct order.

Example 3:

Input: grid = [[1]]\\
Output: [[1]]\\
Explanation:
Diagonals with exactly one element are already in order, so no changes are needed.

Constraints:
\begin{verbatim}
grid.length == grid[i].length == n
1 <= n <= 10
-10^5 <= grid[i][j] <= 10^5
\end{verbatim}
\#\#\# Format: You will use the following starter code to write the solution to the problem and enclose your code within delimiters.
\begin{verbatim}
```python
class Solution:
    def sortMatrix(self, grid: List[List[int]]) -> List[List[int]]:    
``` 
\end{verbatim}

\#\#\# Answer: (use the provided format with backticks)
 and put the final answer in \begin{verbatim}\boxed{}.<|Assistant|><think>\end{verbatim}

\vspace{4pt}
\textbf{Length of reasoning:} 1154 tokens

\vspace{4pt}
\textbf{Model reasoning:}\\
To solve the problem of sorting the diagonals of an n×n matrix as required, we first clarify the core task: sort diagonals in the bottom-left triangle (including the main diagonal) in non-increasing order and those in the top-right triangle in non-decreasing order, then return the modified matrix.

First, we need to identify how to group elements into diagonals. A key observation is that each diagonal in the matrix can be uniquely defined by the difference 
\[
d = \text{row index }(i) - \text{column index }(j).
\]
For an n×n matrix, \(d\) ranges from \(-(n-1)\) (minimum, e.g., \((0, n-1)\) where \(i-j = -n+1\)) to \(n-1\) (maximum, e.g., \((n-1, 0)\) where \(i-j = n-1\)). We further classify diagonals based 

\end{tcolorbox}
\end{figure*}

\begin{figure*}[t!]
\begin{tcolorbox}[colback=gray!5!white,colframe=black!75!black]

on \(d\): if \(d \ge 0\), the diagonal belongs to the bottom-left triangle (including the main diagonal where \(d=0\)); if \(d \textless 0\), it belongs to the top-right triangle. For example, in a 3×3 matrix, \(d=1\) corresponds to the diagonal with positions \((1,0)\) and \((2,1)\), while \(d=-1\) corresponds to \((0,1)\) and \((1,2)\).

The solution follows three core steps: grouping, sorting, and backfilling. First, we traverse the entire matrix to group elements by their \(d\) value. We use a dictionary where the key is \(d\) and the value is a list of \((i,j)\) positions belonging to that diagonal. This ensures we collect all elements of each diagonal for subsequent processing. Next, for each diagonal group, we extract the elements and sort them according to the rule: non-increasing order if \(d \ge 0\) (bottom-left triangle) and non-decreasing order if \(d \textless 0\) (top-right triangle). Finally, we place the sorted elements back into their original positions in the matrix—since we collected positions in the order of increasing row index, we maintain the diagonal’s direction when backfilling.

In code implementation, we first create a deep copy of the input matrix to avoid modifying the original data, which adheres to the principle of side-effect-free functions. We then use \texttt{defaultdict} from the \texttt{collections} module to efficiently group positions by \(d\). For each \(d\) in the dictionary, we extract the elements from the copied matrix, sort them as required, and iterate over the stored positions to assign the sorted elements back. Single-element diagonals (e.g., \(d=2\) in a \(3\times 3\) matrix, corresponding to \((2,0)\)) require no special handling, as sorting a single element has no effect.

\hl{Example 1 illustrates the validity of this approach.} Given the input matrix
\[
[[1,7,3],[9,8,2],[4,5,6]],
\]
we process each diagonal: for \(d=0\) (main diagonal), the elements \([1,8,6]\) are sorted non-increasingly to \([8,6,1]\), which are assigned to positions \((0,0)\), \((1,1)\), and \((2,2)\); for \(d=1\), the elements \([9,5]\) remain unchanged after non-increasing sorting; for \(d=-1\), the elements \([7,2]\) are sorted non-decreasingly to \([2,7]\), assigned to \((0,1)\) and \((1,2)\). The resulting matrix is
\[
[[8,2,3],[9,6,7],[4,5,1]],
\]
which matches the expected output.

\hl{Example 2 further verifies the solution.} For the input matrix
\[
[[0,1],[1,2]],
\]
the main diagonal \((d=0)\) with elements \([0,2]\) is sorted non-increasingly to \([2,0]\), assigned to \((0,0)\) and \((1,1)\); diagonals \(d=1\) (position \((1,0)\)) and \(d=-1\) (position \((0,1)\)) are single-element and unchanged. The final output is
\[
[[2,1],[1,0]],
\]
consistent with the example.

The complete code is as follows:

\begin{verbatim}
from collections import defaultdict
from typing import List

class Solution:
    def sortMatrix(self, grid: List[List[int]]) -> List[List[int]]:
        n = len(grid)
        new_grid = [row.copy() for row in grid]  #Deep copy to preserve input
        diagonals = defaultdict(list)
        
        # Group positions by d = i - j
        for i in range(n):
\end{verbatim}
\end{tcolorbox}
\end{figure*}

\begin{figure*}[t!]
\begin{tcolorbox}[colback=gray!5!white,colframe=black!75!black]
\begin{verbatim}
            for j in range(n):
                d = i - j
                diagonals[d].append((i, j))
        
        # Sort each diagonal group and backfill
        for d in diagonals:
            positions = diagonals[d]
            elements = [new_grid[i][j] for i, j in positions]
            # Sort non-increasing for bottom-left, non-decreasing for top-right
            elements.sort(reverse=True) if d >= 0 else elements.sort()
            # Assign sorted elements back to original positions
            for idx, (i, j) in enumerate(positions):
                new_grid[i][j] = elements
                [idx]
        return new_grid
\end{verbatim}

The core strengths of this approach lie in its clarity and efficiency: grouping by \(d = i - j\) accurately identifies all diagonals, ensuring correct application of sorting rules; deep copying preserves the input matrix; and backfilling in the original position order maintains the diagonal structure. The time complexity is \(O(n^2 \log k)\), where \(k\) is the maximum length of a diagonal (up to \(n\)), which is efficient for n×n matrices.

\vspace{4pt}
\textbf{Final Answer:} \\[4px]
Answer:
To solve this problem, we need to sort the diagonals of a given square matrix in a specific manner. The diagonals in the bottom-left triangle (including the middle diagonal) should be sorted in non-increasing order, while the diagonals in the top-right triangle should be sorted in non-decreasing order.

\#\#\# Approach
1. **Identify Diagonals**: Each diagonal in the matrix can be identified by the difference between the row and column indices (i - j). This difference, referred to as `d`, helps group elements into their respective diagonals.
2. **Group Elements by Diagonals**: For each element in the matrix, compute the diagonal identifier `d` and group the elements based on this identifier.
3. **Sort Diagonals**: For each diagonal group:
   - If the diagonal is part of the bottom-left triangle (i.e., `d \textgreater= 0`), sort the elements in non-increasing order.
   - If the diagonal is part of the top-right triangle (i.e., `d \textless 0`), sort the elements in non-decreasing order.
4. **Reconstruct the Matrix**: After sorting the elements of each diagonal, place them back into their respective positions in the matrix.

\#\#\# Solution Code
\begin{verbatim}
from typing import List
from collections import defaultdict

class Solution:
    def sortMatrix(self, grid: List[List[int]]) -> List[List[int]]:
        n = len(grid)
        new_grid = [row.copy() for row in grid]
        diagonals = defaultdict(list)
        
        for i in range(n):
            for j in range(n):
                d = i - j
                diagonals[d].append((i, j))
        
        for d in diagonals:
\end{verbatim}
\end{tcolorbox}
\end{figure*}

\begin{figure*}[t!]
\begin{tcolorbox}[colback=gray!5!white,colframe=black!75!black]
\begin{verbatim}
            positions = diagonals[d]
            elements = [new_grid[i][j] for (i, j) in positions]
            if d >= 0:
                elements.sort(reverse=True)
            else:
                elements.sort()
            for idx, (i, j) in enumerate(positions):
                new_grid[i][j] = elements[idx]
        
        return new_grid
\end{verbatim}

\[
\boxed{\parbox{1\textwidth}{\centering The ones related to the circulation of the electric field and the divergence of the magnetic field.}}
\]

\end{tcolorbox}

\textbf{Case Analysis.} 
\vspace{3px}
\newline
\begin{minipage}[t]{0.485\textwidth} 
        Considering the unique nature of code reasoning—where correctness cannot be directly verified during the reasoning process and the reasoning must continue until a complete code snippet is generated—the overall reasoning length is inevitably prolonged. Nevertheless, even under such constraints, the optimized reasoning method still exhibits rich example reference behavior (as highlighted in the boxed regions). Such references, semantically grounded in prior examples, effectively reduce the need for self-verification and redundant exploratory steps, thereby enhancing the overall efficiency of the reasoning process.
    \end{minipage}
    \hfill 
  
    \begin{minipage}[t]{0.485\textwidth}
    \end{minipage}

\end{figure*}
\label{tab:}

\end{document}